\documentclass{article}

\usepackage{graphicx}
\usepackage{xcolor}
\usepackage{booktabs}
\usepackage{amsmath}
\usepackage{amssymb}
\usepackage{ifthen}
\usepackage{geometry}
\usepackage{authblk}
\usepackage{endnotes}
\usepackage{hyperref}

\usepackage{url,hyperref,lineno,microtype}

\usepackage{array}
\usepackage{algorithm}
\usepackage{algpseudocode}
\usepackage{tikz}
\usetikzlibrary{positioning,calc,arrows.meta}

\usepackage{natbib}
\setcitestyle{authoryear,round,aysep={}}

\title{From the Fluency Fallacy to the Micro-to-Macro Validity Gap: Opportunities and Pitfalls of LLMs in Social Simulation\thanks{This manuscript has been submitted for peer review to Frontiers in Artificial Intelligence.}}

\author[1]{Patrick Taillandier}
\author[2]{Jean Daniel Zucker}
\author[2]{Arnaud Grignard}
\author[5]{Benoit Gaudou}
\author[4]{Nghi Quang Huynh}
\author[1]{Haojia Kong}
\author[2,3]{Alexis Drogoul}

\affil[1]{UR MIAT, University of Toulouse, INRAE, Castanet-Tolosan, France}
\affil[2]{UMI UMMISCO, IRD, Sorbonne University, Bondy, France}
\affil[3]{IJL ACROSS, IRD, University of Thuyloi, Hanoi, Vietnam}
\affil[4]{CICT, Can Tho University, Can Tho, Vietnam}
\affil[5]{UMR IRIT, University Toulouse Capitole, Toulouse, France}

\begin{document}

\maketitle

\begin{abstract}
The integration of Large Language Models (LLMs) into social simulation has generated considerable enthusiasm, but also raises substantial methodological and epistemological challenges. This critical review examines the use of LLMs as cognitive or decision-making components of simulated agents from a computational social science perspective. Rather than treating the psychological evaluation of LLMs as separate from simulation mechanics, we argue that their behavioral and epistemic limitations can become systemic vulnerabilities when scaled to multi-agent societies. We first map the rapidly evolving landscape of LLM-driven platforms, ranging from small narrative sandboxes to population-scale and spatially structured simulations. We then develop a unified critical framework for analyzing the Micro-to-Macro Validity Gap: the propagation, nonlinear amplification, or convergent contraction of micro-level limitations—including hallucinations, stochastic inconsistency, representational biases, and alignment effects—into agent-level distortions (such as average persona convergence) and macro-level risks (such as artificial consensus and behavioral homogenization), while exposing evaluators to epistemic vulnerabilities such as the Fluency Fallacy and automation bias. We identify contexts in which LLM-based agents offer genuine operational value, including serious games, participatory environments, and exploratory modeling, while distinguishing these uses from confirmatory research with unconstrained models and precise social forecasting. Finally, we examine theory-driven hybrid architectures that embed LLMs within explicit, mechanistic, and reproducible Agent-Based modeling (ABM) frameworks. We argue that such architectures offer a promising but not sufficient path toward improving epistemic control: their validity depends on multi-level evaluation of environmental dynamics, individual behavior, cross-level interactions, and aggregate outcomes, and they remain vulnerable to the risk of physics washing.
\end{abstract}

\noindent 
\textbf{keywords:} Large Language Models, Social Simulation, Agent-Based Modelling, Human behaviour modelling, Epistemology, Computational Social Science
\vspace{1em} 

\section{Introduction}
Large Language Models (LLMs) have emerged as a transformative class of deep learning architectures capable of generating human-like text across a wide range of domains. Built upon the transformer architecture, which employs parallelized attention mechanisms rather than sequential processing, these models are pre-trained on massive corpora comprising web data, encyclopedic content, and code repositories \citep{Vaswani2017Attention,Devlin2019BERT,Brown2020LLMs}. Through unsupervised or self-supervised learning, LLMs acquire syntactic, semantic, and pragmatic language competencies, enabling them to perform a broad array of downstream tasks such as question answering, summarization, translation, and code generation \citep{wang2024factuality}.

Importantly, the magnitude of these training corpora means that LLMs have effectively absorbed an immense statistical imprint of human language use. However, this imprint is far from universally representative; from the outset, the training data exhibits profound structural imbalances, meaning that certain languages and cultures—predominantly Western and English-speaking—are significantly overrepresented compared to others. This goes far beyond learning grammar or syntax: LLMs encode traces of how people reason, argue, empathize, hesitate, and make decisions — all embedded in the linguistic fabric of their training data. Subsequently, these models are fine-tuned via instruction tuning, a supervised process in which models learn to follow natural language commands across a wide range of tasks. This is typically achieved using curated instruction datasets \citep{sanh2021multitask,wei2021finetuned}, sometimes augmented with human feedback to better align model outputs with user intent \citep{ouyang2022training}.

Given this capability to emulate human-like reasoning patterns, LLMs have naturally attracted significant attention within the Agent-Based modeling (ABM) community. However, as their integration into this field accelerates, it is crucial to distinguish between two emerging uses of Large Language Models in computational social science. The first involves using LLMs as coding assistants to facilitate the technical implementation of simulation models (e.g., generating NetLogo or GAMA code). The second, which is the sole focus of this paper, is the use of LLMs as the internal decision-making architecture of the agents themselves—often termed Generative Agents \citep{Park2023GenerativeAgents}. We examine how LLMs can replace or augment traditional rule-based cognition to simulate human-like decision-making, social interaction, and reasoning, rather than how they assist in the software engineering process.

Focusing specifically on this role as an internal decision engine—defined here as the computational mechanism that processes perceptions, maintains memory states, and formulates goal-directed actions, tasks traditionally governed by symbolic architectures such as BDI (Belief–Desire–Intention) or SOAR—a critical area of investigation concerns whether LLM outputs can authentically replicate human behavior.

One classical benchmark in this domain is the Turing Test, proposed by Alan Turing in 1950 as the imitation game, which evaluates whether a machine can produce conversational behavior indistinguishable from that of a human interlocutor \citep{JonesBergen2026Turing}. Recent empirical findings suggest that state-of-the-art LLMs, notably Llama-3.1 and GPT-4.5, have demonstrated convincing performance on a three-party version of this test. In these evaluations, Llama-3.1 was identified as the human participant in 56\% of cases, compared with 73\% for GPT-4.5 \citep{JonesBergen2026Turing}. By applying a standardized three-party design, these results mark a clear departure from classical conversational architectures and rule-based chatbots, which had historically only achieved partial success under less constrained, two-party conditions. These findings underscore the impressive ability of modern LLMs to adopt complex linguistic styles and behavioral nuances, particularly when guided by carefully crafted persona prompts, a form of behavioral prompt engineering that remains a highly precise and non-trivial technique requiring expert design and task sensitivity.

The apparent success of LLMs in the Turing Test has naturally prompted speculation about their suitability for simulating human agents within computational social systems \citep{Park2023GenerativeAgents,anthis2025llmsocialsimulationspromising,KirkebyHinrup2025}. However, the capacity of LLMs to produce believable human-like dialogue does not establish construct validity or mechanistic fidelity as a model of human decision-making. Rather, their performance reflects statistical pattern recognition, i.e., the capacity to reproduce linguistic structures without representing the causal cognitive states of human decision-makers \citep{JonesBergen2026Turing}. Within the philosophy of mind, \citet{dennett1987intentional} characterized the intentional stance as an operational, predictive strategy wherein an observer treats an entity as an intentional agent governed by beliefs and desires, without necessitating that the underlying computational machinery instantiates those internal mental states. While this stance is pragmatic for human-computer interaction, scientific agent-based modeling demands structural and construct validity: mistaking functional behavioral plausibility for mechanistic fidelity risks producing simulations that fail under counterfactual conditions. This distinction between functional imitation and authentic comprehension is particularly salient for social simulation. If LLMs merely replicate surface-level features of human behavior, their utility for generating novel insights into social processes, rather than reproducing known patterns, is fundamentally limited. Moreover, their use as proxies for human agents in scientific studies raises questions about external validity, representational fidelity, and psychological realism. These concerns constitute a central theme of this paper and inform our critical assessment of LLMs as generative agents in artificial societies.

\textbf{Epistemological Stance: A Stratified Typology of Valid Uses.} To avoid ambivalence between categorical critique and uncritical adoption, this review articulates an explicit, four-tier epistemological boundary governing LLM deployment in social simulation:

\begin{enumerate}
\item \textit{Interactive Environments and Serious Games (High Operational Value)}: Generative agents excel where narrative plausibility, contextual engagement, and qualitative role-playing take precedence over mechanistic empirical correspondence, provided factual hallucination guardrails are maintained.
\item \textit{Exploratory modeling (Conditionally Defensible)}: LLMs function legitimately as qualitative reasoning pumps to explore generative assumptions and formulate candidate hypotheses, but only if simulations meet minimum validity criteria (qualitative pattern stability across seeds, invariance under prompt paraphrasing, cross-model portability, and empirical decoupling/falsifiability).
\item \textit{Explanatory and Confirmatory Research (Restricted to Theory-Driven Hybrids)}: Investigating causal mechanisms requires hybrid architectures that embed generative components as delimited subroutines within mechanistic, inspectable ABM engines. Epistemic validity here demands multi-level empirical falsification and demonstrable explanatory gain over minimal, non-generative heuristics to preclude physics washing.
\item \textit{Precise Quantitative Point-Prediction (Methodologically Unsupported)}: Direct empirical forecasting (e.g., fine-grained electoral prediction or precise policy impact estimation) via unconstrained next-token prediction architectures currently exceeds the methodological guarantees of generative models due to prompt sensitivity, stochastic drift, and latent distributional skews.
\end{enumerate}

Unlike previous surveys \citep{Gao2024Survey} and early opportunity mappings \citep{Gurcan2024Challenges}, which primarily organize the technical possibilities and application areas of LLM-augmented simulation, this review focuses on how limitations observed at the level of individual LLM outputs can affect the validity of multi-agent simulations. We argue that the psychological and linguistic evaluation of LLMs cannot be considered independently of the simulation mechanisms through which their outputs are transformed into actions, interactions, and aggregate dynamics.

The review proceeds as follows. Section 2 describes the literature-selection strategy. Section 3 maps the emerging landscape of LLM-based social simulation platforms and identifies recurring architectural patterns, including memory, reflection, planning, communication, and population specialization. Section 4 develops a unified critical framework for the Micro-to-Macro Validity Gap. It examines how hallucinations, stochastic inconsistency, representational biases, alignment effects, and computational constraints may be amplified at the population level, producing risks such as the Fluency Fallacy, convergence toward an average persona, automation bias, and weak longitudinal reproducibility. Section 4 also distinguishes application contexts in which LLM-based agents may provide operational value—such as serious games, participatory environments, exploratory modeling, and narrative interpretation—from applications requiring stronger explanatory or predictive validity. Importantly, we clarify that for off-the-shelf, unconstrained foundation models, confirmatory scientific inference and precise quantitative point-predictions (such as electoral forecasting) currently exceed the methodological guarantees of generative architectures. While hybrid frameworks offer essential guardrails, hybridization serves to bound and contain these epistemic uncertainties within explicit state spaces rather than eliminating them entirely. Determining whether this boundary reflects an intrinsic limitation of next-token prediction or a contingent constraint of current models remains an open empirical question that systematic simulation benchmarks must resolve.

Finally, Section 5 examines theory-driven hybrid architectures as a promising, but not sufficient, methodological direction. It discusses alternative coupling modes, including LLM-as-Subroutine and LLM-as-Policy, as well as the integration of behavioral theories and smaller language models into explicit ABM frameworks. We argue that hybridization can improve environmental grounding, computational control, and the explicitness of behavioral assumptions, but does not by itself establish cognitive or social validity. Hybrid systems therefore require multi-level validation of environmental dynamics, individual decisions, interactions between model components, and aggregate outcomes. The section concludes by identifying physics washing as a central open challenge: the possibility that a formally rigorous simulation environment may lend unwarranted credibility to an insufficiently validated behavioral layer.

\section{Methodology for Literature Selection}
\label{sec:methodology}
To structure this critical review, literature was assembled through a two-stage screening procedure combining a structured database search with targeted exploratory retrieval.

First, a structured query was executed in Scopus on 21 August 2026 restricted to records from 2023 to 2026 using the Boolean expression:
\begin{quote}
\texttt{TITLE-ABS-KEY (("Large Language Model*" OR "Generative Agent*")}\\
\texttt{AND ("Social Simulation" OR "Agent-Based"))}
\end{quote}

This search retrieved 1,266 records (comprising 738 conference papers, 388 journal articles, 55 reviews, and 85 other contributions). Removal of duplicate entries yielded 1,258 unique Scopus records. Second, given the rapid growth of preprint literature and cross-disciplinary intersections with cognitive psychology, an exploratory discovery step was conducted using Gemini 3 Pro (November 2025 release) alongside continuous monitoring of preprint repositories (arXiv, PsyArXiv). This channel served strictly as an index discovery heuristic—not as an evidential source—and surfaced 64 candidate contributions. All candidate items suggested by exploratory retrieval were independently verified against publisher repositories; unverified citations were excluded.

The combined pool of 1,322 records was evaluated against explicit inclusion and exclusion criteria. Studies were included if they: (i) employed LLMs as internal cognitive, deliberative, or decision-making engines for simulated agents rather than auxiliary tools; (ii) investigated social interactions, collective dynamics, or population-level emergence; and (iii) presented verifiable implementations or empirical evaluations. Conversely, studies were excluded if: (i) LLMs were utilized solely as programming or code-generation assistants (e.g., generating NetLogo or GAMA code); (ii) models addressed purely technical multi-agent optimization without behavioral, social, or cognitive depth; or (iii) simulation mechanics were restricted to static question-answering without dynamic agent interactions.

During the title and abstract screening phase, 1,185 records were excluded (1,127 from Scopus and 58 from exploratory retrieval). The remaining 137 publications underwent full-text assessment, resulting in the exclusion of 63 additional Scopus records applying the aforementioned criteria (specifically: purely technical software focus, absence of interactive multi-agent dynamics, or lack of behavioral decision depth). This yielded an analyzed corpus of 74 contemporary studies (2023--2026), comprising 68 empirical and architectural implementations identified via Scopus alongside 6 verified preprint contributions surfaced through exploratory monitoring. This core corpus is complemented by 20 foundational references that provide theoretical and epistemological grounding in agent-based modeling, cognitive architectures, and complex systems (yielding 94 references cited in total).

Screening and qualitative categorization were conducted by the lead investigator following a predetermined, structured protocol, with borderline eligibility cases resolved through consensus discussion among the co-authors. However, several methodological limitations must be acknowledged: (i) title/abstract and full-text screening was performed by a single primary investigator without formal inter-rater reliability metrics (such as Cohen's $\kappa$); (ii) while the core Boolean search in Scopus was designed to capture foundational agent-based and generative simulation contributions, it represents a single curated database query; and (iii) the supplementary exploratory discovery step conducted via Gemini 3 Pro serves as an informal heuristic that is not fully reproducible across time due to model updates and stochastic generation. To ensure research auditability despite these inherent constraints, the complete retrieval queries, date parameters, and operational exclusion criteria were documented and held invariant across screening stages.

\section{The State of LLM-Based Social Simulation: A Transversal Analysis}
\label{sec:state_of_art}

The application of LLMs to social simulation builds upon a long trajectory of integrating machine learning into multi-agent systems \citep{drogoul1998methodological}. However, rather than viewing contemporary platforms merely as independent software packages, an epistemological assessment requires analyzing them through their underlying architectural choices and the systemic tensions they introduce. This transversal perspective makes it possible to compare systems that differ substantially in scale and domain but rely on recurring design choices for memory, planning, interaction, population synthesis, and validation.

To map this landscape, Table~\ref{tab:platforms_comparison} provides a comparative synthesis of prominent platforms across scale, application domain, empirical validation strategy, peer-review publication status, and identified methodological vulnerabilities. Beyond these platform-level implementations, current research reveals recurrent architectural strategies that attempt to operationalize social agency through language models. The following analysis describes these strategies and the trade-offs they create between contextual flexibility, behavioral traceability, environmental grounding, computational tractability, and empirical validation. These trade-offs motivate the micro-to-macro validity gap examined in Section~\ref{sec:validity_gap}.

\begin{table*}[ht]
\centering
\caption{Comparison of state-of-the-art LLM-based Social Simulation Platforms, detailing scale, target domain, validation methodology, publication status, and core methodological vulnerabilities.}
\label{tab:platforms_comparison}
\footnotesize
\setlength{\tabcolsep}{3.5pt}
\renewcommand{\arraystretch}{1.15}
\begin{tabular}{p{0.14\textwidth} p{0.08\textwidth} p{0.13\textwidth} p{0.14\textwidth} p{0.15\textwidth} p{0.30\textwidth}}
\toprule
\textbf{Platform} & \textbf{Scale} & \textbf{Domain} & \textbf{Validation Strategy} & \textbf{Status} & \textbf{Identified Methodological Vulnerabilities} \\
\midrule

\textit{Generative Agents} \citep{Park2023GenerativeAgents} 
& 25 agents 
& General interaction 
& Face validity, emergent coherence 
& Peer-reviewed Conference (ACM UIST)
& Vulnerable to evaluator projection (Fluency Fallacy); lacks empirical benchmark comparison, counterfactual testing, or bounded ground truth. \\

\textit{AgentSociety} \citep{Piao2025AgentSociety}
& 10,000+
& Urban \& social dynamics
& Survey matching (GSS, WVS)
& Preprint (arXiv:2502.08691)
& Aggregate statistical convergence relies on a shared base model prior, risking systemic sycophancy and unverified micro-level decision validity. \\

\textit{GenSim} \citep{Tang2025GenSim}
& Tens of thousands
& General social simulation
& behavioral replication
& Peer-reviewed Conference (NAACL Demo)
& Prone to long-horizon memory degradation, context truncation drift, and high inference costs that preclude extensive parameter sweeps. \\

\textit{$S^3$} \citep{gao2025s3socialnetworksimulationlarge} 
& Variable 
& Social networks 
& Population empirical data 
& Preprint (arXiv:2307.14984)
& Purely linguistic state-transition loop: absence of ground-truth state conservation risks spurious belief propagation insulated from institutional friction. \\

\textit{AgentTorch} \citep{chopra2023agenttorch} 
& Millions 
& General (GPU-accelerated) 
& Statistical robustness 
& Peer-reviewed Workshop (ALOE)
& Stationary distilled policies cannot adapt to feedback, risking frozen micro-rules that mask adaptation-driven macro-dynamics.\\

\textit{SocioVerse} \citep{zhang2025socioverseworldmodelsocial} 
& 10 million 
& Policy testing 
& Empirical distributions 
& Preprint (arXiv:2504.10157)
& Profile-diversity illusion: demographic heterogeneity at initialization risks collapsing into generic normative behaviors during prolonged interaction. \\

\textit{NetLogo Toolchain} \citep{jimenez2025multi}
& Variable
& Multi-agent ABM
& Hybrid grounding
& Peer-reviewed Journal (Front. Artif. Intell.)
& Reactive spatial actuation lacks explicit deliberative cognitive architecture, risking ungrounded state drift in complex socio-institutional environments. \\

\textit{Land System Models} \citep{esd-16-423-2025} 
& Regional 
& Environment 
& Scenario comparison 
& Peer-reviewed Journal (Earth Syst. Dynam.)
& Potential equifinality: aggregate land-use patterns may align with observed trends despite uncertain institutional decision heuristics. \\

\textit{TrajLLM} \citep{Ju2025TrajLLM} 
& Scalable 
& Human trajectory 
& Trajectory comparisons 
& Peer-reviewed Companion (ACM WebConf)
& Disconnect between discrete high-level textual goals and continuous spatial movement; vulnerable to spatial incoherence over long horizons. \\

\textit{Elicitron} \citep{Ataei2025Elicitron} 
& Micro 
& Design engineering 
& Expert evaluation 
& Peer-reviewed Journal (ASME JCISE)
& Rigid task prompting restricts generative freedom, confounding true emergent coordination with scripted task fulfillment. \\
\bottomrule
\end{tabular}
\end{table*}

\subsection{Platform Paradigms and the Scaling Trade-off}

Current platforms instantiate three broad methodological paradigms, reflecting distinct compromises between narrative expressivity, demographic scale, and environmental fidelity:

\begin{enumerate}
    \item \textbf{Narrative and Sandbox Paradigms:} Foundational frameworks prioritize open-ended social interaction, character role-playing, and emergent storytelling. \textit{Generative Agents} \citep{Park2023GenerativeAgents}, whose demonstration environment is commonly known as \textit{Smallville}, instantiates twenty-five agents in a sandbox environment where they maintain daily routines, form relationships, retrieve memories, and generate plans. General-purpose environments such as \textit{GenSim} \citep{Tang2025GenSim} extend this orientation through modular abstractions for social routines, memory orchestration, scheduling, and large-scale interaction. These architectures offer rich qualitative interaction and face plausibility, but remain exposed to temporal inconsistency and memory-management challenges over extended runs, particularly when identity, long-term state, and spatial reasoning depend heavily on retrieved textual context rather than explicit simulation state.
    \item \textbf{Population-Scale and Demographic Paradigms:} A second family emphasizes demographic coverage, population synthesis, and large-scale behavioral mirroring. \textit{AgentSociety} \citep{Piao2025AgentSociety} simulates urban and social dynamics with more than 10,000 LLM-based agents, incorporating mobility, social interaction, and economic activity. \textit{SocioVerse} \citep{zhang2025socioverseworldmodelsocial} uses a pool of profiles associated with 10 million real-world users to construct a heterogeneous synthetic population to explore policy interventions. In social-network simulation, \textit{$S^3$} \citep{gao2025s3socialnetworksimulationlarge} models the propagation of information, attitudes, and emotions, while related work investigates opinion dynamics and polarization in networks of LLM-based agents \citep{chuang2024simulatingopiniondynamicsnetworks}. These approaches increase the scale and empirical diversity of initial conditions. However, demographic diversity in initialization does not by itself establish sustained behavioral diversity in the resulting simulation. Foundation-model priors and alignment procedures may still encourage convergence toward socially normative responses, a possibility that requires direct empirical evaluation rather than being inferred from the diversity of input profiles alone \citep{Wu2025LLMBoundary,lacava2025openmodelsclosedminds}.
    \item \textbf{Spatially Constrained and Scalable Paradigms:} A third family prioritizes explicit environmental structure and computational tractability. \textit{GATSim} \citep{LIU2026103234} models urban mobility through agents that formulate daily travel plans under transportation-network constraints. \textit{MobiVerse} \citep{Liu2025MobiVerse} combines LLMs with lightweight domain-specific generators to reduce the cost of city-scale mobility simulation. \textit{TrajLLM} \citep{Ju2025TrajLLM} combines continuous trajectory generation with discrete LLM reasoning, whereas \textit{LLM-AIDSim} \citep{Zhang2025aLLMAIDSim} applies LLM-based agents to influence diffusion within social networks. In specialized socio-technical settings, \textit{Elicitron} \citep{Ataei2025Elicitron} models interactions involved in engineering design-requirement elicitation. Related work also integrates LLMs into established ABM platforms: a NetLogo toolchain connects prompt-driven agents to an explicit simulation environment \citep{jimenez2025multi}, while land-system models explore LLM representations of institutional agency in human-environment systems \citep{esd-16-423-2025}. These approaches increase environmental grounding and scalability, while often restricting the scope or autonomy of language-model-based decision-making.
\end{enumerate}

\subsection{Transversal Architecture Dimensions: Solutions and Structural Limits}
Across these platforms, LLM-based agents are commonly implemented as separate model instances or sessions conditioned through prompt engineering. Their cognitive architecture typically combines memory, reflection or summarization, planning, and communication modules \citep{Park2023GenerativeAgents,wang2024survey}. These modules are often implemented as chained or nested prompts, with external retrieval systems and orchestration layers providing persistent context and regulating agent interactions. Frameworks such as h-ABM \citep{Atchade_2024} make this decomposition explicit by proposing modular cognitive and emotional layers for integration within agent-based simulations.

\subsubsection{Memory Systems: Retrieval-Augmented Memory and Long-Horizon Consistency}

To support episodic recall and long-term context, many generative-agent architectures use external memory stores, vector databases, and embedding-based retrieval pipelines \citep{Park2023GenerativeAgents,Piao2025AgentSociety}. In these designs, agent observations and interactions are stored as textual traces; a subset of those traces is subsequently retrieved according to relevance, recency, or importance and inserted into the context used for reflection, planning, and action generation. Reflection modules may further summarize recent events, extract higher-level intentions, or prompt agents to generate inferences about the mental states of other agents.These mechanisms operationalize selected aspects of episodic and semantic memory and are sometimes explicitly related to multi-store memory models or theory-of-mind constructs \citep{wang2024survey,Calderon2025CognitiveAgents}.

This modular organization resembles classical symbolic cognitive architectures, including SOAR \citep{Laird2012SOAR} and ACT-R \citep{Anderson2004IntegratedTheory}, which distinguish forms of knowledge and organize the relation between memory, goals, and action selection. However, semantic-similarity retrieval does not by itself encode causal relevance, role-specific priority, or durable commitments. Over extended simulations, the selective retrieval, summarization, and reinterpretation of stored traces can therefore complicate the maintenance of coherent identity, goals, and behavioral history.

\subsubsection{Action Planning: In-Context Reasoning and Reproducibility}

Planning modules typically condition LLM outputs on agent-specific goals, environmental affordances, retrieved memories, and social context. In many systems, structured prompt templates or multi-step reasoning procedures are used to translate these inputs into action plans without specifying an explicit utility function or reward model \citep{Junprung2023ABMSurvey}. Other implementations combine LLMs with formal optimization or learning components. For example, LGC-MARL couples an LLM-based planner with graph-structured multi-agent reinforcement learning: the language model decomposes high-level instructions into executable subtasks and proposes collaborative strategies, whereas the reinforcement-learning component selects or optimizes final actions \citep{Jia2025Enhancing}.

This design enables flexible, context-sensitive action generation, but decisions may remain sensitive to prompt formulation, retrieval context, sampling settings, and model version. Semantically similar situations can consequently yield materially different actions, complicating reproducibility, sensitivity analysis, and counterfactual evaluation.

\subsubsection{Agent Persona: Demographic Prompting and behavioral Diversity}

behavioral diversity is commonly introduced through synthetic-population generation methods and agent-specific prompts. Agents may be initialized with demographic attributes, socioeconomic characteristics, beliefs, occupations, cultural backgrounds, or personality descriptors sampled from survey distributions or human belief networks \citep{chapuis2022generation,chuang2024beyond}. Role-specific memory structures and specialized prompt templates may further differentiate communication styles, goals, and decision heuristics.

These techniques increase descriptive variation in agent inputs, but do not by themselves ensure sustained behavioral heterogeneity. Empirical evaluations suggest that models may revert toward generic or socially normative response patterns despite distinct persona prompts \citep{lacava2025openmodelsclosedminds}. Conversely, stronger attempts to force distinctive roles can lead to exaggerated, stereotypical, or toxic caricatures rather than calibrated behavioral variation \citep{Nudo2026Exaggeration}. The challenge is therefore not only to initialize diverse profiles, but also to maintain and validate theoretically meaningful behavioral differences throughout simulation runs.

\subsubsection{Inter-Agent Communication: Natural Language and Information Propagation}

Simulation time-stepping and inter-agent communication are typically regulated by an orchestration layer. Interaction may occur through direct messages, simulated surveys, public broadcasts, shared memory, or blackboard-style routines. 
Platforms such as \textit{Generative Agents} and \textit{AgentSociety} commonly use orchestration layers that sequence perception, memory retrieval, reflection, planning, and action, whereas \textit{GenSim} incorporates structured social routines and dynamic scheduling. At the software-architecture level, \textit{SALLMA} separates operational processes, such as intention formation and task execution, from knowledge-level components including agent profiles, shared memory, and workflows \citep{becattini2025sallma}. Agents may also be connected to external tools, such as search systems, databases, and calculators, to augment particular deliberative steps \citep{wang2024survey}.

These mechanisms enable flexible social interaction, but also create channels through which hallucinated, misleading, or adversarial content may enter subsequent agent contexts and influence later decisions. In interconnected populations, such information can be retrieved, retransmitted, or incorporated into shared state, making the effects of local errors difficult to isolate and evaluate.

\subsubsection{Environmental Grounding: Generative Proposals and Feasibility Constraints}

LLM-based agents typically express intentions and proposed actions in natural language, whereas an agent-based simulation must translate those proposals into state transitions within an explicit environment. In purely text-based architectures, the language model may therefore be responsible both for proposing an action and for implicitly representing whether the action is spatially, physically, institutionally, or economically feasible. Without an external mechanism that verifies preconditions and updates world state, agents may propose actions incompatible with available resources, network topology, environmental conditions, or scenario rules.

Several recent frameworks address this issue by coupling generative decision-making to explicit simulation environments. In mobility and evacuation applications, transportation networks, spatial accessibility, hazards, and resource constraints can determine which plans are executable \citep{LIU2026103234,Yang2026CrowdEvacuation}. In land-system modeling, LLM-based institutional agents are embedded within a land-use model whose environmental and economic dynamics constrain the consequences of proposed policy actions \citep{esd-16-423-2025}. More generally, architectures such as Concordia distinguish between an agent's language-mediated action proposal and an environmental component that interprets the proposal, checks its feasibility, and applies its consequences to the simulated world \citep{vezhnevets2023generativeagentbasedmodelingactions}.

Environmental grounding can therefore improve the internal consistency of state transitions and make feasibility assumptions explicit. It does not, however, establish that the cognitive or social reasoning that produced a feasible action is psychologically valid. This distinction motivates the theory-driven hybrid architectures and multi-level validation requirements discussed in Section~\ref{sec:towards-theory-driven-hybrid-architectures}.

\subsubsection{Model Adaptation and Computational Execution}

Most LLM-based social simulations rely on frozen foundation models used through in-context learning and prompt engineering. Agent profiles, goals, instructions, retrieved memories, and scenario-specific constraints are supplied at inference time, often together with reflection loops and retrieval-augmented memory. Full parameter adaptation through supervised fine-tuning, preference optimization, or related methods remains less common in narrative simulation settings, partly because suitable behavioral data are limited and because adaptation introduces additional computational and methodological requirements. It is nevertheless used in targeted cognitive modeling to improve correspondence with experimental patterns of human behavior \citep{binz2025foundation}.

A separate challenge concerns execution cost. Repeated calls to a large language model can become expensive when many agents interact over long horizons. Scalable approaches may therefore reduce the frequency of online LLM calls, use lighter domain-specific generators, execute compact policy representations, or rely on smaller task-specific language models. Differentiable and GPU-accelerated ABM frameworks such as \textit{AgentTorch} illustrate the broader value of lightweight policy execution at scale \citep{chopra2023agenttorch}. These choices improve tractability, but may alter the degree of contextual adaptation and behavioral variation available during a simulation run.

\subsection{The Validation Landscape: From Face Validity to Multi-Level Evaluation}
Validation remains a central methodological bottleneck for LLM-based social simulation because the field must move beyond plausible interaction toward evidence that agents, mechanisms, and aggregate outcomes correspond to the intended empirical or theoretical target \citep{Larooij2025Validation,Adornetto2025Validation}. Current practice combines empirical benchmarking, human judgment, statistical comparison, sensitivity analysis, and task-specific datasets, but these approaches remain only partially integrated.

\subsubsection{Empirical Validation and Benchmarking}

Some platforms compare simulated outputs with observed social indicators, including survey-derived attitudinal distributions, economic behavior, voter turnout, or network-level diffusion patterns. For example, \textit{AgentSociety} evaluates simulations against social indicators and survey instruments such as the General Social Survey and the World Values Survey \citep{Piao2025AgentSociety}. Other studies reproduce experimental paradigms, including public-goods games or rumor-propagation settings, and assess whether simulated outcomes correspond to established empirical regularities. 

Such benchmarking provides useful external reference points, although agreement at the aggregate level does not by itself demonstrate that the model generates those patterns through valid individual-level mechanisms. This dilemma is well-established in the classical ABM literature under the rubric of equifinality: multiple distinct micro-level rule sets can generate macro-level patterns that are statistically indistinguishable. To address this challenge, agent-based modeling has historically relied on structural validation, multi-criteria calibration, and the Pattern-Oriented modeling (POM) framework \citep{Grimm2005POM}. POM mitigates equifinality by requiring an architecture to simultaneously reproduce multiple observed empirical patterns across distinct organization levels and spatio-temporal scales, thereby filtering out structurally inadequate behavioral sub-models.

However, the integration of LLMs introduces a qualitative departure from this classical framework. In conventional ABMs, candidate micro-mechanisms are formulated as explicit, inspectable equations or rule sets whose parameters can be constrained through sensitivity analysis and parameter-space rejection filtering. In LLM-driven agents, by contrast, the generative decision mechanism is parameterized within opaque, uninspectable foundation model weights conditioned by natural-language prompts. Consequently, equifinality is no longer merely a problem of parameter indeterminacy among explicit heuristics; it becomes an epistemic barrier where structurally flawed cognitive models can generate statistically plausible aggregate distributions while remaining insulated from direct mechanistic falsification.

\subsubsection{Human-in-the-Loop Evaluation}

Expert judgment remains widely used to assess the believability, internal coherence, and sociological plausibility of agent behaviors. Structured expert protocols may improve inter-rater reliability, while crowdsourced evaluation can provide broader assessments of interaction realism or diversity. Human evaluators may also assess whether agents conditioned on survey-based profiles resemble the intended population \citep{zhang2025socioverseworldmodelsocial}. However, these methods are vulnerable to judgment biases, as linguistic fluency, assertiveness, and perceived authority can lead evaluators to overestimate factual or analytical quality \citep{Chen2024JudgeBias}.

\subsubsection{Robustness and Specialized Evaluation}

Validation can also include exploratory model-behavior analysis, statistical correspondence tests, and sensitivity analysis. The latter examines the stability of results under variation in initial conditions, parameter values, random seeds, prompts, or other design choices. For long-horizon generative simulations, longitudinal consistency is particularly important: researchers must assess whether agents preserve relevant commitments and whether collective outcomes remain reproducible under small perturbations \citep{Wu2025LLMBoundary}. Systems such as \textit{$S^3$} compare simulated information, attitude, and emotion propagation against patterns observed in social-network data \citep{gao2025s3socialnetworksimulationlarge}.

\subsubsection{Data Sources and Comparative Benchmarks}

Task-specific benchmarks complement simulation-level validation. SocialIQA evaluates commonsense social reasoning; CrowS-Pairs and Bias in Bios assess particular forms of social and demographic bias; and
the EPITOME battery evaluates a range of theory-of-mind tasks in humans and language models \citep{Sap2019SocialIQA,Nangia2020CrowSPairs,DeArteaga2019BiasInBios,Karra2024EPITOME}. Robustness benchmarks also test whether semantically equivalent inputs produce stable model responses \citep{Zhang2022Consistency}. These resources are informative for evaluating relevant capabilities and failure modes, but they generally assess isolated tasks rather than the long-horizon, interactive feedback processes characteristic of multi-agent social simulation.

\subsection{Synthesis}
In summary, current LLM-based simulation architectures provide increasingly expressive behavioral proxies and new ways to construct social scenarios at scale. At the same time, each design choice introduces trade-offs and potential sources of error that require explicit examination. When these limitations interact across large populations, they may propagate or be amplified, motivating the Micro-to-Macro Validity Gap examined in Section~\ref{sec:validity_gap}.

\section{The Micro-to-Macro Validity Gap: Critical Challenges}
\label{sec:validity_gap}
Moving from the technical frameworks described above to rigorous scientific investigation raises fundamental epistemological challenges. While current platforms demonstrate that large-scale LLM-based simulations are technically feasible, technical feasibility should not be conflated with scientific validity \citep{Lu2024ComplexSystems}.

To conceptualize how limitations at the model level may affect the scientific validity of multi-agent simulations, we define the \textit{Micro-to-Macro Validity Gap} as:

\begin{quote}
\textit{The gap that arises when limitations and artifacts at the LLM level are translated into individual agent behaviors and subsequently propagated through interactions, potentially producing aggregate dynamics that appear socially plausible while lacking sufficient causal, construct, or external validity.}
\end{quote}

As illustrated in Figure~\ref{fig:validity_gap_framework}, this gap can be understood as a four-level propagation framework:
\begin{enumerate}
    \item \textbf{LLM-Level Properties:} Model-level characteristics such as stochastic generation, factual hallucinations, alignment effects, contextual instability, and demographic or cultural biases in training data.
    \item \textbf{Agent-Level Manifestations:} The translation of these properties into agent behaviors, including convergence toward generic personas, generative variability, sycophancy, longitudinal persona drift, and inconsistent social reasoning.
    
    \item \textbf{Interaction-Level Propagation:} Mechanisms through which individual-level artifacts can propagate, reinforce, or transform through social interactions, including information exchange, shared memory, peer influence, and iterative feedback loops.
    
    \item \textbf{Macro-Level Consequences:} Aggregate patterns that may emerge from these interactions, including artificial consensus, behavioral homogenization, spurious polarization, unstable collective patterns, and other potentially distorted macro-dynamics.
\end{enumerate}

The potential consequence of this cascade is an epistemic risk: researchers may draw policy or theoretical conclusions from simulations that exhibit convincing surface realism while providing limited evidence that the underlying generative mechanisms correspond to those operating in the target social system. To structure this critical evaluation, Table~\ref{tab:challenges_taxonomy} summarizes the main failure modes across cognitive, epistemic, and technical dimensions and maps them to potential hybrid design responses discussed in Section~\ref{sec:towards-theory-driven-hybrid-architectures}.

\begin{figure*}[t]
\centering
\begin{tikzpicture}[
    x=1cm, y=1cm,
    box/.style={
        rectangle, rounded corners=4pt,
        draw=black!75, fill=#1,
        line width=0.85pt,
        text width=3.05cm,
        minimum height=5.3cm,
        align=left, inner sep=5pt,
        font=\fontsize{7.5}{9.5}\selectfont
    },
    arrow/.style={
        -{Stealth[length=6pt,width=4.5pt]},
        line width=1.6pt,
        draw=black!85
    },
    transition/.style={
        font=\fontsize{6.5}{7.8}\selectfont\itshape,
        text=black!85,
        align=center,
        inner sep=1pt
    },
    diagbox/.style={
        rectangle, rounded corners=3pt,
        draw=blue!70!black, fill=blue!6,
        line width=0.6pt,
        text width=3.2cm,
        align=center,
        font=\fontsize{6.5}{7.5}\selectfont\bfseries,
        inner sep=3pt
    }
]

\coordinate (c1) at (0,0);
\coordinate (c2) at (4.65,0);
\coordinate (c3) at (9.30,0);
\coordinate (c4) at (13.95,0);

\node[diagbox] at (2.32, 3.45) {Diagnostic: PAS, refusal, frame-break rates};
\node[diagbox] at (6.97, 3.45) {Diagnostic: DHI (variance contraction)};
\node[diagbox] at (11.62, 3.45) {Diagnostic: Null-model ablation (TOST)};

\node[box=red!8] (llm) at (c1) {
    \centering \textbf{\small Level 1}\\[-1pt]
    \textbf{\scriptsize LLM Limitations \& Biases}\par
    \vspace{0.1cm}\hrule\vspace{0.15cm}
    \raggedright
    \begin{itemize}
        \setlength{\itemsep}{2.5pt}
        \setlength{\leftmargin}{7pt}
        \item Stochastic generation
        \item Factual hallucinations
        \item Training-data imbalances
        \item Alignment \& safety tuning
        \item Context-window drift
    \end{itemize}
};

\node[box=orange!10] (agent) at (c2) {
    \centering \textbf{\small Level 2}\\[-1pt]
    \textbf{\scriptsize Agent Manifestations}\par
    \vspace{0.1cm}\hrule\vspace{0.15cm}
    \raggedright
    \begin{itemize}
        \setlength{\itemsep}{2.5pt}
        \setlength{\leftmargin}{7pt}
        \item ``Average persona'' bias
        \item Uncontrolled variability
        \item Sycophancy \& compliance
        \item Temporal persona drift
        \item Generative caricature
    \end{itemize}
};

\node[box=yellow!15] (interaction) at (c3) {
    \centering \textbf{\small Level 3}\\[-1pt]
    \textbf{\scriptsize Interaction Propagation}\par
    \vspace{0.1cm}\hrule\vspace{0.15cm}
    \raggedright
    \begin{itemize}
        \setlength{\itemsep}{2.5pt}
        \setlength{\leftmargin}{7pt}
        \item Inter-agent dialogue
        \item Memory contamination
        \item Conformism \& peer effects
        \item Iterative feedback loops
        \item Endogenous normative drift
    \end{itemize}
};

\node[box=blue!8] (macro) at (c4) {
    \centering \textbf{\small Level 4}\\[-1pt]
    \textbf{\scriptsize Macro Consequences}\par
    \vspace{0.1cm}\hrule\vspace{0.15cm}
    \raggedright
    \begin{itemize}
        \setlength{\itemsep}{2.5pt}
        \setlength{\leftmargin}{7pt}
        \item Artificial consensus
        \item Variance homogenization
        \item Spurious polarization
        \item Path-dependent drift
        \item Unstable macro-equilibria
    \end{itemize}
};

\draw[arrow] ([yshift=1.1cm]llm.east) -- node[transition, below=3pt] {manifests as} ([yshift=1.1cm]agent.west);

\draw[arrow] ([yshift=1.1cm]agent.east) -- node[transition, below=3pt] {propagates\\through} ([yshift=1.1cm]interaction.west);

\draw[arrow] ([yshift=1.1cm]interaction.east) -- node[transition, above=3pt] {\textbf{Amplification:}\\ Cascades, Thresholds\\[3pt] \textbf{Contraction:}\\ Attractor collapse} ([yshift=1.1cm]macro.west);
\draw[-{Stealth[length=5pt,width=3.5pt]}, dashed, line width=1.1pt, draw=orange!80!black]
    ([yshift=0.2cm]interaction.south west) to[out=230, in=310, looseness=1.2]
    node[transition, below=2pt] {feedback on context}
    ([yshift=0.2cm]agent.south east);

\node[
    rectangle, rounded corners=5pt,
    draw=red!75!black, fill=red!12,
    line width=0.9pt, text width=15.8cm,
    align=center, inner sep=6pt
] (risk) at (6.97, -4.40) {
    \textbf{\footnotesize\textcolor{red!70!black}{
        EPISTEMIC RISK: CONFUSING SURFACE LINGUISTIC REALISM WITH CAUSAL VALIDITY
    }}\\[2pt]
    \scriptsize
    \textbf{Evaluator Biases:} Fluency Fallacy (Discrimination Failure) $\;\bullet\;$ Automation Bias (Uncritical Reliance)\\[1pt]
    \textbf{Systemic Consequences:} Emergent Artifacts Mistaken for Empirical Regularities $\;\bullet\;$ Unfalsifiable Policy Prescriptions
};

\draw[-{Stealth[length=4pt,width=3pt]}, dashed, line width=0.9pt, draw=red!65!black]
    (6.97, -3.10) -- (risk.north);

\end{tikzpicture}
\caption{\textbf{The Micro-to-Macro Validity Gap conceptual framework and diagnostic protocol.}
Individual model-level characteristics (Level 1) manifest as behavioral patterns in simulated agents (Level 2). Through interaction networks, shared memory, and conversational feedback (Level 3), individual errors can undergo divergent nonlinear amplification (via information cascades or threshold phase transitions) or convergent contraction (attractor collapse toward dominant model priors), yielding distorted macro-level dynamics (Level 4). Concrete diagnostics ($PAS$, $DHI$, and null-model ablation via TOST) monitor transitions between successive levels. The framework highlights that human evaluators face epistemic vulnerabilities (the Fluency Fallacy and automation bias), risking mistaking superficial narrative coherence for causal empirical validity.}
\label{fig:validity_gap_framework}
\end{figure*}

\begin{table*}[ht]
\centering
\caption{Taxonomy of methodological challenges in LLM-based social simulation and potential hybrid design responses. The final column identifies architectural directions discussed in Section 5; these approaches may improve epistemic control or computational tractability, but do not by themselves establish the validity of the resulting simulation.}
\label{tab:challenges_taxonomy}
\scriptsize                           
\setlength{\tabcolsep}{3.5pt}         
\renewcommand{\arraystretch}{0.92}    

\begin{tabular}{p{0.13\textwidth} p{0.27\textwidth} p{0.28\textwidth} p{0.28\textwidth}}
\toprule
\textbf{Category} & \textbf{Challenge \& Mechanism} & \textbf{Impact on Social Simulation} & \textbf{Potential Hybrid Design Response (Sec. 5)} \\
\midrule

\textbf{Cognitive \& behavioral} & 
\textbf{The Average Persona Risk} \newline Statistical regression toward the mean driven by pre-training corpus distributions \citep{Wang2025AveragePersona,Wu2025LLMBoundary}. & 
\textit{Replicant Effect}: Loss of population diversity; erasure of minority distributions and extreme behaviors. & 
\textbf{Theory-Grounded Agent States}: Represent selected demographic distributions, beliefs, and behavioral rules explicitly within the ABM rather than relying solely on prompt-defined personas. \\
\cmidrule{2-4}
 & 
 
 \textbf{Alignment-Induced Sycophancy} \newline Post-training preference optimization (RLHF/DPO) inadvertently inducing compliant, conflict-averse responses \citep{sharma2023towards,Li2025ActionsSpeak}. & 
\textit{Synthetic Harmony}: Agents refuse realistic conflict, partisan polarization, or non-normative decisions, yielding artificial consensus. & 
\textbf{Domain-Specific / SLMs without conversational preference tuning}: Deploy local, task-specialized language models free from post-hoc safety conversational tuning, combined with theory-guided prompt scaffolding. \\
\cmidrule{2-4}
 & 
\textbf{Generative Exaggeration} \newline Tendency to caricature specific roles when prompted away from the mean \citep{Nudo2026Exaggeration}. & 
\textit{Toxic Caricature}: Agents exhibit stereotypical or extreme behaviors rather than nuanced human decision-making. & 
\textbf{Theory and Environment Constraints}: Combine environmental feasibility checks with explicit behavioral assumptions and validation criteria to identify implausible or stereotyped outputs. \\
\midrule

\textbf{Methodological \& Epistemic} & 
\textbf{The Fluency Fallacy} \newline Evaluator discrimination failure: high linguistic plausibility masking underlying causal invalidity \citep{Steyvers2024DiscriminationGap}. &
\textit{Surface Realism}: Simulations appear valid to human evaluators but lack causal veridicality. & 
\textbf{Structural Evaluation Safeguards (Sec. 4.2)}: Deploy (i) ablation of narrative traces to evaluate symbolic logs directly, (ii) deterministic ledger and invariant verification, and (iii) adversarial counterfactual auditing. \\

\cmidrule{2-4}
 & 
\textbf{Epistemic Opacity} \newline Implicit behavioral rules hidden in billions of parameters \citep{bommasani2022opportunitiesrisksfoundationmodels}. & 
\textit{Unverifiable Emergence}: Inability to trace unexpected macroscopic behaviors back to explicit agent decision rules. & 
\textbf{Traceable Hybrid Architecture}: Assign explicit state variables, environmental checks, and specified action-selection mechanisms to the ABM, while restricting LLM use to clearly delimited functions. \\
\cmidrule{2-4}
 & 
\textbf{Data Scarcity Wall} \newline Lack of training data for non-WEIRD populations \citep{Atari2023WhichHumans,Suhaib2024Perils}. & 
\textit{Cultural Erasure}: Inability to simulate specific marginalized groups via pure fine-tuning. & 
\textbf{Theory-Driven Alignment}: Use explicit behavioral theories to structure assumptions and expose uncertainties where representative empirical data are sparse. \\
\midrule

\textbf{Technical \& Operational} & 
\textbf{Temporal Instability} \newline Context window limits and stochastic drift over long horizons \citep{Wu2025LLMBoundary}. & 
\textit{Longitudinal Incoherence}: Agents forget their history or change personality during long simulations. & 
\textbf{Externalized State Management}: Store selected identities, commitments, social ties, and memory summaries in explicit ABM-managed state rather than relying solely on the LLM context window. \\
\cmidrule{2-4}
 & 
\textbf{Computational Bottleneck} \newline High inference costs preventing large-scale iterative runs. & 
\textit{No Sensitivity Analysis}: Inability to perform rigorous statistical validation (1000+ runs). & 
\textbf{Offline Policies and Lightweight Models}: Distil, approximate, or execute selected policies offline, and use task-specific SLMs where appropriate to reduce repeated inference costs. \\
\cmidrule{2-4}
 & 
 
 \textbf{Adversarial \& Misuse Risks} \newline Susceptibility to prompt injection and emergent normative drift observed in exploratory testbeds \citep{akkil2026emergenceworldplatformevaluating}. & 
\textit{Systemic Pollution}: Compromised behavior spreads through shared memory, distorting the collective reasoning of the population. & 
\textbf{Constrained Interaction Protocols (Sec. 4.7)}: Define typed, permissioned, monitored, and auditable communication channels to prevent unconstrained natural-language prompt injection from entering shared agent memory.\\
\bottomrule
\end{tabular}
\end{table*}

\subsection{Formalizing Amplification and Diagnostic Evaluation}
\label{sec:formalizing_amplification}

To complement the qualitative analysis, the following quantities and protocols are formulated as conceptual diagnostic proposals to guide future experimental validation, rather than empirical estimates computed within this review. A critical distinction must be maintained between linear error propagation and nonlinear distortion across the four levels. Let $\epsilon$ denote the magnitude of a controlled micro-level perturbation (e.g., the fraction of corrupted memory entries or injected hallucinated assertions) and $E(T;\epsilon)$ the divergence at horizon $T$ between perturbed and unperturbed macro-trajectories across paired seeds. We formally distinguish three operational regimes:
\begin{enumerate}
    \item \textbf{Linear propagation.} $E(T;\epsilon) \propto \epsilon$ (scaling exponent $\alpha=1$) and underlying transition rules remain unaltered. In this baseline regime, micro-level artifacts enter collective state additively without altering the surrounding decision landscape.
    
    \item \textbf{Divergent amplification (cascades, threshold transitions).} Two distinct criteria must be reported separately, as they are not mathematically equivalent: 
    \begin{itemize}
        \item \emph{Amplitude nonlinearity}: $E(T;\epsilon) \propto \epsilon^{\alpha}$ with $\alpha > 1$, estimated by log--log regression across multiple perturbation magnitudes $\epsilon$;
        \item \emph{Temporal instability}: a positive growth rate $\lambda = \lim_{T\to\infty}\tfrac{1}{T}\log E(T;\epsilon) > 0$ (a finite-horizon divergence rate analogous to a Lyapunov exponent, corresponding in the linearized state formulation $e_{t+1} = \mathbf{G}e_t + \xi_t$ to a spectral radius $\rho(\mathbf{G}) > 1$). Formal estimation of a true Lyapunov exponent would require additional assumptions concerning state representation, coupling, stochasticity, and local dynamical smoothness.
    \end{itemize}
    A linear system with $\rho(\mathbf{G}) > 1$ is dynamically unstable yet maintains $\alpha = 1$, whereas a threshold network mechanism can yield $\alpha > 1$ without sustained exponential growth. Divergent amplification in the strict sense requires $\alpha > 1$ and/or $\lambda > 0$.
    
    \item \textbf{Convergent contraction (attractor collapse).} Let $d_t = \tfrac1N \sum_i \lVert x_{i,t} - x^\ast \rVert$ denote the mean distance of agent states to the base model's default normative prior $x^\ast$. Attractor collapse corresponds to $d_t \to 0$ (in the linearization around $x^\ast$, $\rho(\mathbf{G}) < 1$) and $DHI(t) \to 1$. This is contraction rather than divergent amplification: the error relative to the \emph{human target} does not diverge, but converges to a persistent non-zero bias (the distance between $x^\ast$ and the empirical target distribution), explaining why it systematically eludes conventional temporal instability diagnostics.
\end{enumerate}

Outside these explicit dynamic and contraction regimes, the four-level framework operates as a structured diagnostic heuristic. To empirically detect these regimes and monitor error propagation across successive stages of the modeling pipeline, targeted diagnostic metrics should be deployed across each inter-level transition:
\begin{itemize}
 \item \textbf{Level 1 $\rightarrow$ Level 2 (Model to Agent):} Track the composite Persona Adherence Score ($PAS$), defined multiplicatively across prompt iterations to prevent double-counting:
\begin{equation}
PAS = (1 - R_{\text{refusal}})(1 - R_{\text{frame-break}})\,R_{\text{adherence}\mid\text{valid}} \in [0, 1]
\end{equation}
where $R_{\text{refusal}} \in [0, 1]$ indexes refusal rates driven by alignment guardrails, $R_{\text{frame-break}}$ records the frequency of meta-linguistic persona breaks (e.g., reverting to \textit{``As an AI language model...''}), and $R_{\text{adherence}\mid\text{valid}} \in [0, 1]$ measures semantic and behavioral alignment with assigned persona descriptors evaluated strictly over valid, non-refused responses. Importantly, $PAS$ measures fidelity to the assigned persona, not correspondence with a human population; $\tau_{1/2}(PAS)$ is therefore a persona-fidelity horizon, to be complemented by an independently defined distance to the empirical target.
    
    \item \textbf{Level 2 $\rightarrow$ Level 3 (Agent to Interaction):} We propose a Dynamic Homogenization Index ($DHI$), formalizing the cumulative contraction of between-agent attitudinal variance across interaction steps $t$:
    \begin{equation}
    DHI(t) = 1 - \frac{\mathrm{Var}(X_t)}{\mathrm{Var}(X_0)}, \quad \text{with } \mathrm{Var}(X_0) > 0
    \end{equation}
    where $X_t$ denotes continuous behavioral choices or projection embeddings along specific ideological or value dimensions. A step-wise rate of contraction can be monitored via $\Delta DHI(t) = DHI(t) - DHI(t-1)$. $DHI$ is dimension-specific and should be interpreted only relative to a pre-specified behavioral or attitudinal projection. Negative values indicate increased dispersion rather than a measurement failure.
    
    \item \textbf{Level 3 $\rightarrow$ Level 4 (Interaction to Macro):} Conduct interaction-ablation tests using a distribution-matched null model ($\mathcal{N}_1$). Researchers simulate the identical macro-structural interaction topology, replacing generative LLM cognitive engines with non-interactive agents whose static micro-level choice probabilities are empirically matched to the marginal distributions produced by the LLM agents. If macro-level patterns (e.g., consensus formation, polarization regimes) under the generative model and the $\mathcal{N}_1$ null model exhibit statistical equivalence under two one-sided tests (TOST) within an a priori equivalence margin calibrated to inter-seed variance, this strongly suggests that complex inter-agent generative deliberation provides no identifiable causal explanatory power over static choice distributions in the evaluated regime.
\end{itemize}

\subsection{Epistemic Risks: Hallucinations and the Fluency Fallacy}
\label{sec:epistemic}

At the micro-level, evaluating LLMs as agent decision mechanisms reveals a fragile correspondence with human psychological processes \citep{Strachan2024Theory, Ullman2023LLMsFail, Karra2024EPITOME}. In addition, LLMs are fundamentally constrained by stochastic inconsistency and hallucinations—generating baseless outputs while maintaining a coherent linguistic structure \citep{wang2024factuality}. 

When these micro-level flaws are embedded in multi-agent populations, they generate or amplify an epistemic vulnerability that we term the \textit{Fluency Fallacy}. We conceptualize this phenomenon fundamentally as an \textit{evaluator discrimination failure}: the systematic inability of human evaluators to distinguish causally valid from invalid agent decisions when masked by fluent narrative justifications. While intuitive, this concept directly builds upon established traditions in cognitive psychology and human-computer interaction regarding processing fluency heuristics, epistemic trust, and automation bias \citep{Cummings2012AutomationBias, Steyvers2024DiscriminationGap}. In cognitive science, the fluency heuristic posits that subjective ease of processing serves as an implicit proxy for truth and validity \citep{AlterOppenheimer2009Fluency, ReberSchwarz1999Truth}; similarly, in HCI authority bias research, perceived articulateness systematically inflates an automated system's perceived competence. As discussed by \citet{Li2026GhostOfBehaviorism}, this concern reflects a broader methodological problem and a recurring neo-behaviorist dilemma in computational psychology: researchers frequently mistake surface-level linguistic performance for internal cognitive or social validity.

\textbf{Operational Definition and Measurement.} Operationally, while a simple heuristic expression of the Fluency Fallacy measures the elevated subjective plausibility granted to objectively invalid decisions, $\mathbb{E}[S_{\text{plausibility}}(\text{Trace}) \mid V_{\text{causal}}(\text{Decision}) = 0]$, this metric alone cannot separate evaluator leniency from true cognitive illusion. To formally capture this diagnostic breakdown across continuous populations of evaluation traces, we propose an operational index of the Fluency Fallacy based on the Area Under the Receiver Operating Characteristic Curve ($\text{AUC}$) of subjective plausibility scores $S \in [0, 1]$ predicting objective validity $V \in \{0, 1\}$ (evaluated against explicit ground-truth rules, ledger conservation, or counterfactual stability within the ABM environment):
\begin{equation}
FF_{\text{AUC}} = 1 - \text{AUC}\left(S_{\text{plausibility}} \rightarrow V_{\text{causal}}\right)
\end{equation}

Under this formulation, $\text{AUC}(S \rightarrow V) \approx 0.50$ would indicate that evaluators perform no better than chance in distinguishing valid from invalid decisions under the specified evaluation protocol, while $\text{AUC} < 0.50$ indicates that subjective plausibility is systematically misleading (in the absence of score-direction reversal or coding artifacts).

\textbf{Trigger Conditions and Vulnerable Populations.} This fallacy is most pronounced under three conditions: (i) high semantic complexity with sparse factual feedback (e.g., policy deliberations where outcome delays prevent immediate error detection); (ii) long conversational horizons where local linguistic coherence remains intact while global causal continuity drifts; and (iii) scenarios lacking strict physical or ledger constraints. We hypothesize that domain experts without simulation modeling background (who focus primarily on qualitative persona realism) may be particularly susceptible to this evaluative trap. Furthermore, this vulnerability may be amplified in individuals exhibiting high cognitive dependence on automated conversational agents, a construct formalized by the LLM dependence scale of \citet{Li2025LLMDependenceScale}.

\textbf{Methodological Safeguards.} To mitigate the Fluency Fallacy, simulation frameworks must incorporate structural safeguards that decouple behavioral evaluation from surface-level linguistic narratives:
\begin{enumerate}
    \item \textit{Ablation of Narrative Traces during Validation}: Evaluating simulated agents based strictly on symbolic, numerical, and state-transition logs rather than generated discursive justifications. Evaluators inspect state vectors directly—such as resource transfers, spatial coordinates, time-budget allocations, or voting tallies—using conventional agent-based validation methods (e.g., transition matrices, trajectory clustering, and time-series divergence). By removing the linguistic framing, this blind-evaluation protocol prevents human evaluators from projecting rational intentionality onto causally inconsistent behavior.
    
    \item \textit{Ledger and Deterministic Environmental Verification}: Enforcing independent conservation laws, budget balances, and physical invariants directly within the ABM engine rather than accepting textual declarations at face value. As demonstrated in preliminary evaluations of multi-agent environments such as \textit{Emergence World} \citep{akkil2026emergenceworldplatformevaluating}, generative agents can maintain conversational plausibility and strictly adhere to dialogue constraints while simultaneously issuing baseless or fabricated claims regarding resource possession. A deterministic ledger layer verifies whether proposed transactions or material actions satisfy feasibility preconditions prior to state commitment, thereby preventing ungrounded hallucinations from polluting the persistent simulation state.
    
    \item \textit{Adversarial Counterfactual Auditing}: Systematically perturbing antecedent parameters to test the stability and falsifiability of the underlying decision heuristic. Language models can generate post-hoc rationalizations that fit an observed scenario by retrieving standard linguistic patterns rather than executing a causal decision process. By executing micro-interventions on localized input variables (e.g., inverting price incentives, shifting regulatory constraints, or modifying social cues) and comparing agent responses across parallel seeds, researchers can verify whether decisions exhibit theoretically consistent directional shifts or degenerate into arbitrary stochastic variance.
\end{enumerate}

\subsection{Representational Risks: Cognitive Bias, Average Persona, and Alignment}
\label{sec:representational_risks}

\subsubsection{Pre-training Distributional Biases and the Average Persona}
\label{sec:average_persona}
While foundational studies frequently invoke the general concept of Western, Educated, Industrialized, Rich, and Democratic (WEIRD) bias \citep{Atari2023WhichHumans, Suhaib2024Perils}, recent empirical benchmarks have begun to systematically quantify both the magnitude and directionality of these representational skews across standardized value frameworks and demographic strata.
Using large-scale survey instruments such as the Pew Research surveys (OpinionQA), \citet{Santurkar2023WhoseOpinions} demonstrated that instruction-tuned and RLHF-aligned models exhibit marked baseline skews: default responses align predominantly with liberal, college-educated, and secular demographics in the United States. \citet{Durmus2023GlobalOpinionQA} also reported systematic differences between model outputs and subjective-opinion distributions across countries and demographic groups, highlighting the difficulty of treating general-purpose models as culturally representative populations. Similarly, psychometric probing across Hofstede's cultural dimensions and Moral Foundations Theory reveals that pre-training corpora under-represent traditional, conservative, and collectivistic moral matrices, skewing disproportionately toward individualistic fairness and harm reduction \citep{Atari2023WhichHumans}.

In a social simulation context, these pre-training distributional imbalances generate systemic homogenization. While classical ABMs intentionally employ simplified, homogeneous agents (the KISS paradigm \citep{axelrod1997advancing}), LLM-driven architectures aspire to the KIDS (Keep It Descriptive, Stupid) paradigm \citep{edmonds2005kids} to model richly diversified human populations. However, the underlying statistical mechanics of next-token prediction actively undermine this goal, triggering an \textit{Average Persona Convergence} \citep{Wang2025AveragePersona} whose severity is modulated by identifiable conditions:
\begin{enumerate}
    \item \textit{Conditioning Depth and Prompting Regimes}: Behavioral convergence toward the population mean is particularly pronounced under minimal demographic conditioning, such as relying solely on coarse socioeconomic labels (e.g., \textit{``You are a 55-year-old rural factory worker''}). In low-context configurations, model outputs consistently revert to default statistical priors. While elaborate narrative backstories introduce lexical diversity, empirical evaluations indicate that agents frequently converge on homogeneous moral and operational decisions during extended multi-agent interactions, a dynamic termed the \textit{Replicant Effect} \citep[cf.][]{Wu2025LLMBoundary, Li2025ActionsSpeak}.
    \item \textit{Asymmetric Impact across Populations}: Behavioral homogenization varies systematically across demographic strata \citep{Wang2025AveragePersona}. Sub-populations that are extensively documented in digital English-language corpora (such as urban, educated Western cohorts) typically retain greater behavioral variance under persona conditioning. In contrast, under-represented demographic groups—such as non-Western or traditional rural communities—are particularly vulnerable to representational attenuation, where localized decision-making patterns are systematically replaced by the model's dominant normative baselines \citep{Wang2025AveragePersona, Atari2023WhichHumans}.
\end{enumerate}

Furthermore, attempting to force models away from this statistical average persona via aggressive prompt engineering frequently triggers \textit{Generative Exaggeration}: rather than calibrating authentic psychological variance, agents drift into toxic caricatures and stereotypical extremes, replacing statistical homogenization with sociological parody \citep{Nudo2026Exaggeration}.

\subsubsection{Safety Alignment, Preference Tuning, and Modeler Dependence}
\label{sec:alignment_effects}
A critical distinction must be drawn between statistical regression to the mean arising from pre-training corpus distributions and the active behavioral compression induced by post-training alignment procedures (e.g., supervised instruction tuning, RLHF, and DPO). 

First, as documented by \citet{lacava2025openmodelsclosedminds} across multiple open-weights model architectures, supervised instruction tuning and preference optimization consistently constrict downstream behavioral variance, causing diverse base models to converge toward restricted, normative behavioral repertoires. 

Second, while deliberate safety alignment is intended to suppress toxic, harmful, or illegal outputs, it inadvertently fosters an emergent \textit{sycophancy bias}—an unprompted tendency to agree with user cues and avoid confrontational positions \citep{sharma2023towards}. When deployed in social simulation, safety-tuned agents frequently suppress realistic partisan, conflictual, or non-normative behaviors \citep{sharma2023towards, Li2025ActionsSpeak}, defaulting instead to polite and artificially compliant consensus. As demonstrated by \citet{Li2025ActionsSpeak}, these post-training alignment artifacts do not merely affect surface textual dialogue but directly skew the operational choices and downstream actions agents execute within the simulated environment.

This post-training compression interacts directly with modeler trust and reliance. As highlighted by the psychometric framework of \citet{Li2025LLMDependenceScale}, researchers can develop cognitive dependence on LLMs, which may attenuate scrutiny of model-generated content. In simulation design, an over-reliance on automated outputs risks reducing critical inspection during prompt engineering and validation: investigators may be more inclined to interpret aligned, cooperative dialogue as plausible human social dynamics rather than as an artifact of safety tuning. Consequently, post-training alignment artifacts and unchecked evaluator reliance can mutually reinforce an artificial consensus, potentially obscuring more contentious, polarized, or heterogeneous population dynamics.

\subsection{Interpretability Risks: Opacity and Automation Bias}
\label{sec:opacity}
The opacity of LLMs can also contribute to automation bias \citep{Cummings2012AutomationBias}, whereby users tend to place undue trust in automated systems and accept their outputs with limited scrutiny. This risk is amplified by the ability of LLMs to produce plausible but unsupported or incorrect explanations while maintaining a high level of linguistic confidence, making erroneous outputs difficult to identify without external verification \citep{Simhi2024TrustMeImWrong}.

This issue contrasts with the more explicit behavioral specification traditionally used in agent-based modeling. In conventional ABMs, researchers define behavioral rules, assumptions, and decision mechanisms explicitly, and these can be inspected and calibrated against empirical observations. In LLM-driven agents, by contrast, much of the behavioral specification is implicit in the model's parameters and learned representations \citep{bommasani2022opportunitiesrisksfoundationmodels, Larooij2025Validation}. Consequently, when an LLM-based agent produces an unexpected decision-for example, by deviating from an expected economic strategy or displaying a socially biased response-it can be difficult to identify the mechanisms responsible for that behavior. Researchers may therefore struggle to determine whether an observed pattern reflects a meaningful emergent phenomenon at the population level or an artifact of the model's stochasticity, training data, or latent representations.

This limitation is particularly consequential when simulations are used to inform policy or other high-stakes decisions. Decision-support systems in such contexts require not only predictive or behavioral performance, but also sufficient transparency, robustness, and accountability to support the interpretation and scrutiny of their outputs \citep{Ju2024Sense}. When the computational mechanisms underlying simulated agent decisions remain largely opaque, biases or artifacts originating from the underlying LLM may be propagated through the simulation without being readily identifiable. This can undermine the interpretability and epistemic validity of simulation-based analyses and, consequently, the reliability of policy recommendations derived from them \citep{echterhoff2024cognitive}.

\subsection{Temporal Dynamics: Degradation Regimes and Characteristic Timescales}
\label{sec:temporal_dynamics}

In classical ABMs, long-term behaviors, asymptotic tendencies, and system sensitivity are typically characterized through extensive Monte Carlo simulations and structured sensitivity analyses over invariant, well-defined state spaces. In LLM-driven simulations, by contrast, evaluating temporal stability faces two interrelated barriers: the high computational cost of running sufficient replications, and the non-stationary nature of agent cognition, which is continuously mediated through accumulating textual history, selective memory retrieval, and recursive summarization \citep{Wu2025LLMBoundary}. We hypothesize that this temporal degradation is not uniform but unfolds conceptually across three successive phases:

\begin{enumerate}
    \item \textit{Early-Phase In-Context Coherence}: During initial interaction cycles, agents closely adhere to their prompted personas and operational guidelines. Decisions remain directly anchored in system-level prompts, with behavioral variance reflecting stochastic sampling rather than structural drift.
    
    \item \textit{Mid-Phase Retrieval and Contextual Drift}: As interaction histories exceed operational context windows, architectures rely on embedding-based retrieval (RAG) and periodic reflection mechanisms \citep{Park2023GenerativeAgents, Tang2025GenSim}. Because semantic similarity does not track temporal hierarchies or causal necessity, agents retrieve fragmented anecdotes. Behavioral consistency progressively weakens: initial commitments are reinterpreted, and contextual personas experience uncalibrated drift.
    
    \item \textit{Late-Phase Recursive Compaction and Attractor Collapse}: Over extended horizons, agents generate secondary reflections based on previously summarized memories and prior ungrounded statements. Under unconstrained conversational dynamics, this recursive compression leads to compounding information loss, where diverse individual personas risk collapsing toward the baseline normative prior of the underlying foundation model \citep[cf.][]{Wu2025LLMBoundary}, driving aggregate trajectories toward artificial consensus or unreproducible fluctuations.
\end{enumerate}

\textbf{Characteristic Timescales and Operational Half-Life.} To transition from a qualitative description to an empirical benchmark, we formalize the operational reliability timescale via the \textit{persona adherence half-life}, $\tau_{1/2}(PAS)$. Building upon the Persona Adherence Score defined in Section~\ref{sec:formalizing_amplification}, $\tau_{1/2}(PAS)$ denotes the characteristic simulation horizon (measured in discrete time-steps $t$ or interaction rounds) at which persona fidelity decays to half its baseline level:
\begin{equation}
\tau_{1/2}(PAS) = \min \left\{ t \;\middle|\; \mathbb{E}[PAS(t)] \le \frac{1}{2}\mathbb{E}[PAS(0)] \right\}
\end{equation}
We hypothesize that in purely conversational, prompt-based architectures, this characteristic timescale is tightly bounded by the effective uncompressed context capacity before recursive summarization dilutes behavioral fidelity \citep{Wu2025LLMBoundary}. Simulations extending beyond $\tau_{1/2}(PAS)$ are expected to lose longitudinal construct validity unless stabilized by hybrid architectures (Section~\ref{sec:towards-theory-driven-hybrid-architectures}) that externalize persistent state variables, commitments, and structural constraints directly within the ABM environment.

\subsection{Computational Constraints and Scalability}
Although LLMs may offer logistical advantages over large-scale human experiments, they introduce substantial computational constraints compared with traditional Agent-Based Models (ABMs). Their resource requirements can be considerable, particularly for local deployment, which may require substantial GPU resources, or for cloud-based inference, where large-scale simulations generate a high volume of API requests \citep{Costa2025GamifiedSystems, xia2024understanding}. When thousands of narrative agents interact repeatedly over extended simulation periods, inference latency and computational or API costs can quickly become significant, limiting the practical scalability of such approaches.

This computational overhead has methodological implications beyond runtime and infrastructure requirements. Many forms of analysis used to assess the robustness of stochastic social simulations rely on repeated model executions under different parameter settings and initial conditions. Such experiments are essential for characterizing parameter sensitivity, exploring boundary conditions, and quantifying the variability of simulation outcomes \citep{Piao2025AgentSociety, Wu2025LLMBoundary}. While lightweight rule-based agents can often support extensive parameter sweeps and replications on standard CPUs, the substantially higher cost of LLM inference can make comparable experimental designs impractical. Researchers may therefore face a trade-off between the behavioral richness afforded by LLM-based agents and the number of simulations that can be feasibly performed.

Several strategies can mitigate this constraint, including reducing the number of agents or interactions, limiting the frequency of LLM calls, and replacing online inference with offline or distilled policies \citep{chopra2023agenttorch}. However, these strategies may themselves affect the properties being studied. Reducing population size can constrain the analysis of macro-level patterns and emergent phenomena, while replacing online inference with distilled policies may alter the behavioral variability and contextual adaptation of the original LLM. Consequently, computational scalability is not simply an engineering consideration: it can directly constrain the experimental designs required to rigorously analyze and validate LLM-based social simulations.

\subsection{Behavioral Perturbation, Adversarial Dynamics, and Scientific Instability}
\label{sec:adversarial}

Beyond technical and computational limitations, LLM-driven simulations introduce acute risks when agents or their environments are exposed to communicative or adversarial perturbations. In classical rule-based ABMs, agent behavioral repertoires are bounded by explicit state-transition logic, meaning that external perturbations induce responses strictly within the parameter space defined by the modeler. In contrast, LLM-based agents operate over an open-ended, context-sensitive behavioral manifold. While this enables expressive social role-play, it creates an attack surface where prompt injection \citep{greshake2023youvesignedforcompromising} and jailbreaking techniques \citep{zou2023universaltransferableadversarialattacks} can manipulate decision heuristics by exploiting conflicts between system-level directives and conversational context. This vulnerability is particularly present for agents conditioned to sustain detailed social personas, as misleading cues embedded in ongoing dialogues become incorporated into persistent context and alter subsequent reasoning \citep{from_individual_to_society}.

\textbf{Systemic Contamination and Endogenous Normative Drift.} When deployed within populations, such vulnerabilities cease to be isolated anomalies and can manifest as systemic dynamics. Recent exploratory evaluations in sandbox testbeds such as \textit{Emergence World} suggest that simulated prompt-injection and phishing events may propagate across agent societies: when agents exchange messages, maintain shared memories, or consult common ledgers, compromised inputs produced by a single agent can be retrieved by neighboring peers, potentially contaminating collective memory and altering population-level behavior \citep{akkil2026emergenceworldplatformevaluating}.

This vulnerability may also manifest without deliberate external manipulation. Preliminary multi-agent experiments indicate that agents exhibiting aligned, cooperative behavior in isolation or within homogeneous groups can undergo endogenous \textit{normative drift} when interacting with agents powered by alternative model architectures \citep{akkil2026emergenceworldplatformevaluating}. While these exploratory findings warrant systematic replication across diverse platforms, they illustrate how peer-to-peer influence and iterative feedback loops can dynamically modify behavioral baselines, showing that safety and behavioral stability cannot be assessed solely through isolated single-agent benchmark testing.

\textbf{Epistemic Consequences: Falsifiability and Path-Dependent Irreproducibility.} These perturbation dynamics carry direct epistemological consequences that distinguish generative simulations from traditional complex systems models:
\begin{enumerate}
    \item \textit{Rupture of Counterfactual Falsifiability}: When an agent adopts non-normative or unpredicted choices following a social exchange, researchers cannot reliably determine whether this represents genuine emergent social behavior or an unconstrained linguistic drift triggered by semantic perturbations in the prompt history. The lack of bounded state invariants leaves behavioral hypotheses largely insulated from rigorous falsification.
    \item \textit{Path-Dependent Irreproducibility}: Because semantic shifts in inter-agent communication enter subsequent context windows, minor conversational fluctuations can trigger discontinuous phase shifts across the agent network. This produces high longitudinal path-dependency, preventing the stable replication of macro-trajectories across parallel stochastic seeds.
\end{enumerate}

Adversarial and conversational sensitivity is therefore not merely a cybersecurity concern, but a structural vulnerability within the Micro-to-Macro Validity Gap: local linguistic instability directly undermines the replicability, interpretability, and scientific defensibility of aggregate simulation findings. This instability further underscores the need for explicitly bounded, theory-driven hybrid architectures (Section~\ref{sec:towards-theory-driven-hybrid-architectures}) capable of regulating inter-agent communication channels.

\subsection{Ethics in Application: Serious Games vs. Predictive Science}

Given the methodological limitations discussed above, it is important to distinguish between application domains in which LLM-based agents can provide practical value and those in which their limitations may compromise scientific validity. In particular, the requirements for an interactive learning environment differ substantially from those of a simulation intended to support empirical claims or quantitative predictions.

\subsubsection{Interactive Environments and Serious Games}

In educational games, training environments, and participatory simulations \citep{taillandier2019participatory}, LLMs can provide substantial benefits. In these settings, the primary objective is often to create engaging, adaptive, and plausible interactions rather than to reproduce social behavior with strict empirical fidelity \citep{Costa2025GamifiedSystems}. LLMs can dynamically adapt dialogue and scenarios to individual users, enabling personalized learning pathways and richer narrative interactions \citep{Liu2025TeachingPlans}. They can also be combined with traditional machine-learning methods to monitor player behavior and identify indicators of cognitive difficulty or disengagement during educational activities \citep{Liu2025DetectStruggle}.

These benefits do not eliminate the need for safeguards. In educational and training contexts, incorrect or misleading outputs can have direct consequences for learning. Unlike a rule-based simulator in which the space of possible outcomes can be explicitly specified and validated, an LLM-based agent may generate plausible but factually incorrect information. Such outputs can contribute to superficial learning or excessive reliance on the system \citep{Zhu2025Superficial}. Biases in the underlying model can introduce additional problems, particularly when agents are expected to portray different social roles or perspectives. For example, \citet{arana2025foundations} report inconsistencies in role-play behavior and the reproduction of stereotypical patterns in personalized educational simulations. Consequently, the use of LLMs in educational or training environments should combine their flexibility with mechanisms for factual grounding, systematic evaluation, and adversarial testing. Retrieval-augmented generation (RAG), constrained prompting, and human oversight can help reduce, although not necessarily eliminate, these risks.
\subsubsection{From Prediction to Exploratory modeling: Minimum Validity Standards}
\label{sec:exploratory_standards}

For social simulations intended as scientific instruments, the epistemic limitations detailed in Sections~\ref{sec:epistemic} and~\ref{sec:opacity} mean that deploying LLM-based agents for direct confirmatory empirical research or fine-grained point-predictions (such as precise electoral forecasting) currently exceeds the methodological guarantees of generative architectures. Given their sensitivity to contextualization, prompt semantics, and stochasticity, establishing a stable, verifiable correspondence with real-world target populations remains a major challenge. Even when supported by architectural scaffolding, generative components in high-stakes explanatory settings should be adopted conditionally: their scientific value depends on demonstrating independent empirical falsifiability and measurable explanatory gains over minimal, non-generative baseline rules.

A more defensible role for LLM-based agents is therefore exploratory modeling \citep{Lu2024ComplexSystems}, where simulations function as computational reasoning pumps to explore the consequences of alternative behavioral assumptions \citep{Wu2025LLMBoundary}. However, treating a model as exploratory cannot serve as an escape clause that exempts it from methodological scrutiny. Unconstrained exploratory claims risk producing behavioral patterns that reflect artifacts of model pre-training or prompt phrasing rather than genuine socio-behavioral dynamics.

This tension is formalized by \citet{LiWu2025PromptDilemma} as the structural dilemma of prompt engineering in generative agent-based modeling: balancing prompt-level control against emergent behavioral freedom. At one extreme, highly prescriptive prompting rigidly predetermines agent actions, reducing the simulation to a scripted demonstration with zero generative autonomy. At the other extreme, unconstrained prompting leaves agents vulnerable to stochastic drift and hallucinations, yielding scientifically vacuous outputs. 

For exploratory simulation to be methodologically defensible, it must satisfy minimum validity criteria:
\begin{enumerate}
    \item \textit{Qualitative Pattern Robustness Across Seeds}: While numerical outputs in generative systems vary, exploratory hypotheses are defensible only if macroscopic qualitative regimes (e.g., segregation thresholds, polarization, or cooperation dynamics) persist reliably across multiple random seeds and temperature settings.
    \item \textit{Prompt-Sensitivity Invariance}: Exploratory insights must not depend on fragile phrasing. As highlighted in architectural and robustness evaluations (Sections~\ref{sec:state_of_art} and~\ref{sec:epistemic}), candidate hypotheses must demonstrate stability under semantic paraphrasing of instructions and persona descriptors.
    \item \textit{Cross-Model and Version Portability}: Emergent dynamics should be evaluated across distinct model families (e.g., contrasting open-weight architectures). An emergent pattern observable in only one proprietary model checkpoint may reflect idiosyncratic alignment artifacts rather than a generalizable social dynamic.
    \item \textit{Empirical Decoupling and Falsifiability}: Exploratory modeling must conclude with structured, testable propositions that can be decoupled from the generative software environment and submitted to independent empirical evaluation (e.g., via behavioral experiments, surveys, or field observations).
\end{enumerate}

\subsubsection{Narrative Generation and Results Interpretation}

A further application of LLMs is to use them not as the decision-making mechanism of simulated agents, but as tools for interpreting simulation outputs. In this approach, LLMs operate either post-hoc or alongside the simulation to transform complex model outputs into qualitative descriptions, scenarios, or narratives that can be more readily examined by researchers and stakeholders (e.g. \citep{arnejo2025automatic}). This can facilitate the interpretation and communication of macro-level patterns by connecting quantitative simulation results with human-readable descriptions of possible trajectories and social dynamics. This use is methodologically distinct from employing an LLM as the primary decision-making mechanism of each simulated agent. Because the underlying simulation remains responsible for generating the quantitative results, the LLM primarily acts as an interface for interpretation and communication. Nevertheless, this separation does not remove the risk of introducing unsupported interpretations. Generated narratives may attribute causal relationships that are not present in the simulation or emphasize patterns that are not statistically supported. LLM-based interpretation should therefore remain grounded in the underlying model outputs, with explicit constraints on the information available to the model and, where possible, systematic verification of generated claims. This separation between simulation and interpretation may offer a comparatively low-risk way of exploiting LLMs’ language capabilities while preserving the transparency and reproducibility of the underlying scientific model.

\section{Towards Theory-Driven Hybrid Architectures}
\label{sec:towards-theory-driven-hybrid-architectures}

The limitations of purely generative approaches suggest that LLMs may be most useful when embedded within structured Agent-Based Models (ABMs) and informed by explicit behavioral or cognitive theories \citep{taillandier2019building, jimenez2025multi}. In such architectures, the ABM provides an explicit representation of the simulated environment—including spatial, physical, institutional, and resource constraints—while the language model supports selected aspects of agent deliberation. This structural separation addresses an essential vulnerability of purely generative frameworks: an LLM may propose an action based on textual context, but the ABM deterministically evaluates whether that action is compatible with the physical state and regulatory invariants of the simulated world. Compatibility with environmental constraints should not, however, be confused with the empirical or construct validity of the generative decision mechanism itself.

Beyond environmental grounding, hybrid systems provide an architectural framework for decomposing human deliberation into intermediate processes informed by cognitive psychology and behavioral economics, rather than relying on monolithic end-to-end prompting. Such scaffolding constrains generative variability, renders behavioral assumptions inspectable, and retains the contextual interpretability of language models.

However, hybridization does not eliminate the stochasticity or representational limits of language models; rather, it bounds them within explicit and inspectable state spaces. By delegating physical invariants, resource ledgers, and institutional rules to the deterministic simulation engine, hybrid designs can prevent some proposed actions from violating explicit environmental or accounting constraints, while restricting language models to modular cognitive subroutines.

\subsection{Coupling Modes: LLM-as-Subroutine vs. LLM-as-Policy}
\label{sec:coupling_modes}

Current integrations of LLMs into simulation engines can broadly be organized into two operational paradigms:

In the \textit{LLM-as-Subroutine} configuration, the ABM engine executes core system mechanics (spatial motion, physical flows, baseline accounting), while the language model is queried synchronously for complex social or qualitative reasoning tasks. This separation maintains deterministic environmental tracking while leveraging contextual deliberation in domains where closed-form mathematical heuristics are unavailable \citep{Yang2026CrowdEvacuation, Shi2026EmpiricalGame}. 

The second configuration, \textit{LLM-as-Policy}, shifts language model computation offline. Rather than executing iterative calls during runtime, model outputs are compiled or distilled into parameterized behavioral policies executed deterministically by the simulation engine \citep{chopra2023agenttorch}. This design drastically reduces computational overhead, permitting extensive parameter sensitivity testing, albeit at the expense of runtime conversational adaptation.

An operational compromise is offered by Small Language Models (SLMs) \citep{Gao2025SmallLanguageModels}. When fine-tuned for specialized decision tasks, localized SLMs run within simulation loops at substantially lower latency, facilitating replication while eliminating external API dependencies. However, as evaluated in our concluding research agenda (Section~\ref{sec:conclusion}), whether task-specialized SLMs can preserve authentic behavioral heterogeneity without succumbing to variance constriction remains an open empirical question that must be validated independently of their computational advantages. Furthermore, hybrid designs externalize persistent state: agent memory, social ties, and material endowments are maintained and updated directly within explicit ABM data structures rather than entrusted to a drifting context window \citep{Park2023GenerativeAgents, Piao2025AgentSociety}.

\subsection{behavioral Alignment with Empirical Data and Theory-Driven Scaffolding}
\label{sec:behavioral_alignment}

A prominent strategy for improving agent realism involves aligning models with empirical psychological datasets \citep{binz2025foundation}. However, empirical alignment faces a fundamental boundary: the availability of high-resolution behavioral data is sparse for many demographic groups and decision contexts, and general-purpose corpora remain skewed toward WEIRD populations \citep{Atari2023WhichHumans, Suhaib2024Perils}. Sole reliance on observed datasets therefore risks reinforcing existing sampling blind spots.

To address this scarcity wall, hybrid architectures introduce explicit theoretical scaffolding. Rather than generating unstructured behavioral outputs, language model queries are constrained by established behavioral frameworks, such as the Theory of Planned behavior (TPB) \citep{ajzen1991theory, scalco2018application}. Under this framework, models estimate intermediate, theory-specified constructs—attitudes, subjective norms, and perceived behavioral control—which are stored as explicit state variables before intention is translated into action. This structural scaffolding renders causal assumptions inspectable, limits generative drift, and creates intermediate targets for empirical calibration.

\subsection{Conceptual Example: Theory-Driven Coupling in Agricultural Irrigation}
\label{sec:case_study}

This subsection outlines a conceptual blueprint to examine how these theoretical scaffolds and coupling mechanisms operate in practice. Specifically, we consider a stylized scenario of agricultural irrigation under drought-induced water restrictions.

While established socio-hydrological platforms such as MAELIA \citep{Therond2014MAELIA} model physical catchment dynamics (streamflow deficits, groundwater storage, crop evapotranspiration) and institutional governance rules (prefectural restriction tiers) deterministically without relying on language models, we examine here a hypothetical hybrid architecture where generative components are embedded within such a grounded environment. In this conceptual implementation, agent deliberation is structured across three sequential phases:

\begin{enumerate}
\item \textbf{Environmental Tracking and Externalized State (ABM Engine):} The simulation engine deterministically computes biophysical balances (e.g., daily soil moisture deficits modeled through mechanistic crop modules) and tracks statutory irrigation quotas and regulatory restriction stages (e.g., partial or total irrigation bans). Each farmer agent maintains an explicit, inspectable state vector outside the language model, including historical crop yields, operating debt, and irrigation withdrawal allowances.
    
\item \textbf{Theory-Structured Cognitive Deliberation (SLM Component):} Rather than deploying an unconstrained prompt (e.g., \textit{``Decide whether to irrigate your maize field today''}), deliberation is structured through the Theory of Planned behavior \citep{ajzen1991theory}. A local Small Language Model (SLM) is queried via modular prompts solely to evaluate intermediate psychological constructs:

\begin{itemize}
\item \textit{Attitude} ($A \in [0, 1]$): Evaluated against expected crop economic loss versus the perceived risk and penalty of non-compliance fines under current restriction alerts.
\item \textit{Subjective Norms} ($SN \in [0, 1]$): Conditioned on the observed compliance or defiance behaviors of neighboring peers within the farmer's social network.
\item \textit{Perceived Behavioral Control} ($PBC \in [0, 1]$): Evaluated strictly as the agent's \textit{subjective self-efficacy} (e.g., perceived ability to implement alternative scheduling or manage bureaucratic restriction exceptions). This subjective construct is decoupled from hard physical constraints to avoid double-counting with subsequent deterministic ledger checks.
\end{itemize}
  
To prevent uninspectable heuristic aggregation by the language model, the SLM is restricted to estimating separate numerical scores for each construct. The ABM engine then computes the behavioral intention score via an explicit, calibrated weighting equation:
\begin{equation}
I = w_A A + w_{SN} SN + w_{PBC} PBC, \quad \text{with } w_A + w_{SN} + w_{PBC} = 1
\end{equation}

The ABM deterministically maps this intention score $I \in [0, 1]$ to an operational irrigation volume proposal $V_{\text{prop}}$:
\begin{equation}
 V_{\text{prop}} = I \cdot V_{\text{agronomic}}
\end{equation}

where $V_{\text{agronomic}}$ denotes the objective daily water requirement estimated mechanistically by the crop-growth module. The linear mapping from intention to volume is itself a behavioral assumption and must be calibrated against metered withdrawals; it is a potential locus of physics washing. The additive intention equation and the linear mapping from intention to water volume are illustrative formalizations rather than direct empirical consequences of the TPB. Their parameters and functional forms would require independent calibration and comparison with alternative specifications.

\item \textbf{Feasibility Verification and Biophysical Feedback (ABM Engine):} The proposed volume $V_{\text{prop}}$ is returned to the ABM engine, which executes deterministic boundary verification: irrigation is capped if the pumping station exceeds legal hourly flow thresholds, if farm pond reserves are exhausted, or if water rights are legally suspended. The verified withdrawal $V_{\text{exec}} \le V_{\text{prop}}$ is subtracted from hydrological catchment storage, updating downstream streamflow and generating biophysical feedback for subsequent simulation steps.
\end{enumerate}
\textbf{Multi-Level Validation Protocol.} This hybrid decomposition enables systematic evaluation across four connected structural levels, mirroring the analytical architecture of the validity gap:
\begin{itemize}
    \item \textit{$V_1$: Biophysical and Environmental Grounding}: Mechanistic hydrological equations (evapotranspiration, river discharge, groundwater depletion) are validated independently against physical benchmark data (e.g., historical hydrological gauging stations and meteorological time series).
    \item \textit{$V_2$: Micro-Level Cognitive Calibration}: Intermediate psychological scores ($A, SN, PBC$) estimated by the SLM are calibrated against agricultural survey datasets, discrete choice experiments, and post-drought qualitative interviews.
    \item \textit{$V_3$: Cross-Level Interaction and Face Validity}: To verify decision heuristics without narrative distortion, stylized decision logs (soil moisture status, alert stage, neighbor decisions, and executed withdrawal) are audited via stakeholder-in-the-loop panels (farmers and watershed managers) using blind operational timelines decoupled from generated linguistic justifications.
    \item \textit{$V_4$: Macro-Level Socio-Hydrological Trajectories}: Aggregate seasonal withdrawal curves and catchment-scale depletion rates are benchmarked against regional water-metering records and satellite observations of crop greenness. It is at this level ($V_4$) that the minimal heuristic substitution test ($\mathcal{N}_2$) is formally evaluated to ensure the generative layer delivers demonstrable explanatory gain over parsimonious baseline rules.
\end{itemize}

By decoupling the biophysical ledger from cognitive reasoning, this framework helps mitigate the risk of physics washing, supporting that individual decisions remain causally traceable and falsifiable without mistaking environmental rigor for behavioral validity.

\subsection{Open Challenges: The Validation Trap and Physics Washing}
\label{sec:physics_washing}

As illustrated by the need for multi-level calibration in socio-environmental systems, hybrid architectures do not eliminate epistemic challenges; rather, they redistribute them across distinct model components. A persistent vulnerability is that explicit and deterministic components of an ABM may create an impression of overall rigor that is not warranted for the underlying behavioral or generative components. We refer to this risk as \textit{physics washing}: an epistemic distortion whereby the mathematical, spatial, or physical precision of the environment confers unwarranted scientific credibility onto an unvalidated or speculative behavioral layer.

\textbf{Conceptual Lineage and Analogy with Greenwashing.} The term draws a deliberate analogy with greenwashing. In greenwashing, superficial environmental claims mask unsustainable operational practices; similarly, in physics washing, rigorous physical conservation laws, high-resolution GIS topographies, or mechanistic equations serve as a methodological veneer that obscures the epistemic opacity of prompt-driven decision heuristics. The risk is particularly acute in interdisciplinary modeling teams, where domain specialists from natural or physical sciences may assume that because the physical layer is calibrated against empirical benchmark datasets, the coupled socio-behavioral dynamics inherit equivalent validity.

\textbf{A Typology of Physics Washing Scenarios.} We propose three recurring conceptual configurations in which physics washing may arise:
\begin{enumerate}
    \item \textit{Spatial and GIS Dressing}: Simulations deploying high-resolution topographic or urban road networks (e.g., in mobility and evacuation modeling) where spatial routing, collision avoidance, and network capacities are modeled with formal accuracy, but destination choice, panic thresholds, or risk perception are governed by uncalibrated conversational prompts.
    \item \textit{Biophysical Interface Masking}: Socio-ecological architectures (e.g., agricultural hydrology, epidemiological models) in which mechanistic physical balances (soil water retention, pathogen transmission matrices) are coupled to LLM agents whose compliance, risk tolerance, or resource-use heuristics are sensitive to unverified prompt phrasing.
    \item \textit{Formal Ledger Camouflage}: Economic or market simulations where double-entry financial ledgers, strict clearinghouses, and conservation of money enforce mechanical accounting validity, concealing the fact that agent bidding, utility estimation, or negotiation strategies fail basic axioms of consumer rationality.
\end{enumerate}

\textbf{Diagnostic Criteria and Auditing Guidelines.} While the safeguards detailed in Section~\ref{sec:epistemic} protect evaluators against deceptive linguistic narratives, auditing a simulation against physics washing requires diagnostics targeted specifically at the structural coupling between physical and behavioral layers:
\begin{enumerate}
    \item \textit{Decoupled Layer Calibration}: Behavioral decision heuristics must be calibrated against independent empirical human data prior to integration into the physical simulation environment. High goodness-of-fit metrics in the biophysical or spatial layer (e.g., matching regional hydrological discharge or traffic network flows) must not be accepted as evidence that the behavioral sub-model is valid.
    
    \item \textit{Cross-Layer Sensitivity Analysis}: Conducting systematic stress tests to evaluate how perturbations propagate across the interface. A scientifically sound hybrid model must demonstrate bidirectional responsiveness: significant biophysical changes (e.g., severe resource scarcity) should induce non-trivial behavioral adjustments, while minor prompt variations must not trigger disproportionate, chaotic phase transitions in the physical environment.

    \item \textit{Explanatory Gain via Minimal Heuristic Substitution}: Re-running the integrated simulation after replacing the generative LLM component with a minimal baseline rule ($\mathcal{N}_2$) calibrated on the same empirical source data (treating uniform random draws strictly as sanity checks rather than competing baselines). If aggregate socio-environmental trajectories remain statistically equivalent under two one-sided tests (TOST) with an a priori equivalence margin accounting for stochastic variance, the generative agent layer provides no identifiable explanatory gain and functions merely as decorative complexity. While decorative generative complexity is not identical to physics washing per se, it becomes an instance of physics washing when the formal rigor of the biophysical engine is leveraged to claim scientific validity for the unvalidated generative layer.

\end{enumerate}

\section{Conclusion}
\label{sec:conclusion}
Large Language Models have demonstrated remarkable capabilities in natural language interaction and achieve strong performance across standardized reasoning tasks \citep{JonesBergen2026Turing, Ullman2023LLMsFail}. However, the literature reviewed here indicates that linguistic fluency should not be taken as evidence of causal cognitive mechanisms. The resulting divergence between verbal coherence and behavioral validity produces the Fluency Fallacy: an evaluator discrimination failure whereby human judges mistake surface-level linguistic plausibility for authentic causal social agency \citep{Steyvers2024DiscriminationGap, Larooij2025Validation}. In computational social science, this distinction is fundamental, as articulate narrative traces frequently mask hallucinated assertions, ungrounded heuristics, and latent demographic skews.

As generative simulation scales from sandbox environments \citep{Park2023GenerativeAgents} toward large populations \citep{Piao2025AgentSociety} and spatially explicit frameworks \citep{LIU2026103234}, these micro-level vulnerabilities become systemic. Individual persona convergence, prompt sensitivity, and training-set distributional biases propagate through inter-agent interactions, amplifying into the Micro-to-Macro Validity Gap: aggregate outcomes can exhibit superficial realism while remaining causally disconnected from the target social system \citep{Wu2025LLMBoundary, lacava2025openmodelsclosedminds}. Furthermore, opacity, non-stationary temporal drift, and high inference costs make the direct deployment of unconstrained LLM agents problematic for confirmatory science and predictive policy analysis.

These limitations do not negate the utility of language models in simulation; rather, they clarify their methodological boundaries. Generative agents provide immediate operational value in interactive learning environments, serious games, and participatory simulations where narrative expressiveness and contextual engagement take precedence over empirical correspondence \citep{Costa2025GamifiedSystems}. Similarly, within exploratory modeling, language models serve as computational reasoning pumps to examine alternative behavioral scenarios and generate qualitative hypotheses under explicit prompt constraints \citep{Lu2024ComplexSystems, LiWu2025PromptDilemma}.

For explanatory research, the evidence reviewed points toward theory-driven hybrid architectures that embed language models within explicit, mechanistic ABM platforms \citep{taillandier2019building, chopra2023agenttorch}. Yet hybridization does not automatically resolve epistemic vulnerability. A formally rigorous environmental engine can enforce spatial, physical, or accounting constraints while leaving the generative behavioral layer uncalibrated, creating the risk of physics washing.

To prevent hybrid modeling from remaining an abstract design concept and anchor it in empirical falsification, we propose three concrete, forward-looking research benchmarks for future evaluation:

\textbf{1. Standardized Counterfactual Benchmarking for Behavioral Fidelity.} 
Evaluating whether persona-conditioned agents exhibit genuine causal reasoning or merely stochastic paraphrasing under situational change. 
\textit{Protocol}: Deploy an open benchmark where simulated agents are subjected to controlled micro-interventions on incentive structures (e.g., changes in taxation, resource costs, or regulatory sanctions) while holding the environmental baseline constant. 
\textit{Success Criterion}: The simulated standardized effect size $d_{\text{sim}}$ preserves the directional sign predicted by behavioral theory across all stochastic seeds and prompt paraphrasings, and falls within a pre-specified equivalence margin $\delta_d$ of the empirical human reference effect size $d_{\text{hum}}$ for the identical intervention.
\textit{Falsifiable Failure Criterion}: Systematic sign reversals across paraphrases, failure to achieve theoretically consistent directional shifts, or inter-seed variance exceeding $\delta_d$ demonstrates an absence of causal grounding, restricting the architecture to exploratory scenario generation.

\textbf{2. Stress-Testing the Decoupling Barrier to Prevent Physics Washing.} 
Establishing whether coupling an LLM to an explicit environmental engine provides demonstrable explanatory gain over non-generative heuristics. 
\textit{Protocol}: Systematic baseline-substitution experiments using an empirically calibrated minimal rule baseline ($\mathcal{N}_2$) within identical biophysical environments (such as the agricultural irrigation setup in Section~\ref{sec:case_study}). To ensure a fair comparison, $\mathcal{N}_2$ must be calibrated on the same empirical source data as the generative layer (e.g., a threshold or fixed-construct TPB decision rule fitted to survey or choice-experiment data, with its number of free parameters explicitly reported), whereas uniform random draws serve strictly as sanity checks rather than competing baselines. Let $\Delta = \mathrm{RMSE}_{\mathcal{N}_2} - \mathrm{RMSE}_{\text{LLM}}$ denote the paired difference in macro-level misfit across seeds, and $\delta > 0$ the minimum substantively meaningful improvement.
\textit{Success Criterion}: A one-sided superiority test rejects $H_0: \Delta \le \delta$, confirming that generative deliberation contributes measurable explanatory value beyond parsimonious rules.
\textit{Falsifiable Failure Criterion}: If two one-sided tests (TOST) establish statistical equivalence within a predefined margin, $|\Delta| < \delta_{\text{eq}}$, the generative component provides no causal explanatory gain and operates merely as decorative complexity—constituting physics washing whenever the formal rigor of the biophysical engine is leveraged to claim unearned validity for the coupled cognitive layer. (Runs failing both superiority and equivalence remain inconclusive).

\textbf{3. Empirical Calibration of Small Language Models without Conversational Preference Tuning.} 
Evaluating whether localized SLMs without conversational preference tuning can serve as calibrated cognitive subroutines free from the consensus bias and conflict suppression induced by safety tuning. 
\textit{Protocol}: Fine-tune open-weights SLMs strictly on domain-specific behavioral datasets (e.g., attitudinal survey batteries, choice-experiment records) bounded by intermediate psychological constructs, explicitly bypassing conversational RLHF/DPO layers. To mitigate post-training variance constriction (Section~\ref{sec:alignment_effects}), models should be fine-tuned on target response distributions rather than deterministic modal answers, sampled stochastically, and verified for baseline distributional calibration prior to multi-agent interaction.
\textit{Success Criterion}: The Wasserstein distance between simulated and empirical human attitude distributions (as well as polarization indices) remains within an equivalence margin bounded by observed between-sample human variability across successive interaction rounds. The Dynamic Homogenization Index ($DHI$) is evaluated as a secondary diagnostic against empirical human trajectories, accommodating settings where human populations naturally polarize ($DHI < 0$).
\textit{Falsifiable Failure Criterion}: If specialized, theory-scaffolded SLMs undergo artificial variance collapse ($DHI \to 1$) or fail to sustain observed empirical heterogeneity, this would provide systematic evidence of an operational boundary in current autoregressive architectures, demarcating where statistical next-token prediction cannot substitute for explicit mechanistic agent modeling.

Addressing these empirical challenges will enable the computational social science community to systematically demarcate where generative agents genuinely enhance simulation science and where classical mechanistic modeling remains methodologically irreplaceable.

\section*{Conflict of Interest Statement}
The authors declare that the research was conducted in the absence of any commercial or financial relationships that could be construed as a potential conflict of interest.

\section*{Author Contributions}
PT conceived the original structure and wrote the first draft of the manuscript. JDZ, AG, BG, NQH, HK and AD substantially contributed to the conceptualization, provided critical feedback, and improved the manuscript. All authors discussed the concepts, read, and approved the final manuscript.

\section*{Funding}
This work was supported by the French National Research Agency (Agence Nationale de la Recherche, ANR) under the project RL-MAT.

\section*{Data Availability Statement}
No new datasets were generated or analyzed in this study. All literature analyzed is referenced within the article.

\bibliographystyle{Frontiers-Harvard} 
\bibliography{references}


\end{document}